\documentclass[journal]{IEEEtran}

\ifCLASSINFOpdf
\else
\fi
\usepackage{graphicx}
\usepackage{booktabs}
\usepackage{amsmath}
\usepackage{rotating}
\usepackage[pagebackref=true,breaklinks=true,colorlinks,linkcolor=blue,citecolor=blue]{hyperref}
\usepackage{color}
\usepackage{times}
\usepackage{epsfig}
\usepackage{graphicx}
\usepackage{amssymb}
\usepackage{multirow}
\usepackage{algorithm}
\usepackage{algorithmic}
\usepackage{array}
\usepackage{cite}
\newcommand{\RN}[1]{%
  \textup{\uppercase\expandafter{\romannumeral#1}}%
}

\usepackage[utf8]{inputenc}
\usepackage[T1]{fontenc}
\usepackage{textcomp}
\usepackage[table]{xcolor}
\usepackage{threeparttable}
\usepackage{makecell} 
\usepackage{dcolumn}
\usepackage{siunitx}
\begin{document}

\title{SMC-NCA: Semantic-guided Multi-level Contrast for Semi-supervised Temporal Action Segmentation}
\author{Feixiang Zhou, Zheheng Jiang, Huiyu Zhou and Xuelong Li, \textit{Fellow, IEEE} 
\thanks{F. Zhou Z. Jiang and H. Zhou are with School of Computing and Mathematical Sciences, University of Leicester, United Kingdom. E-mail: \{fz64;zj87;hz143\}@leicester.ac.uk. H. Zhou is the corresponding author. The supplementary material is available at \href{https://github.com/FeixiangZhou/SMC-NCA}{Supp.}, which includes additional experiments and explanations.}
\thanks{X. Li is with School of Artificial Intelligence, Optics and Electronics (iOPEN), Northwestern Polytechnical University, Xi’an 710072, P.R. China. E-mail: li@nwpu.edu.cn.}

}

\maketitle

\begin{abstract}
Semi-supervised temporal action segmentation (SS-TAS) aims to perform frame-wise classification in long untrimmed videos, where only a fraction of videos in the training set have labels. Recent studies have shown the potential of contrastive learning in unsupervised representation learning using unlabelled data. However, learning the representation of each frame by unsupervised contrastive learning for action segmentation remains an open and challenging problem. In this paper, we propose a novel Semantic-guided Multi-level Contrast scheme with a Neighbourhood-Consistency-Aware unit (SMC-NCA) to extract strong frame-wise representations for SS-TAS. Specifically, for representation learning, SMC is first used to explore intra- and inter-information variations in a unified and contrastive way, based on action-specific semantic information and temporal information highlighting relations between actions. Then, the NCA module, which is responsible for enforcing spatial consistency between neighbourhoods centered at different frames to alleviate over-segmentation issues, works alongside SMC for semi-supervised learning (SSL). Our SMC outperforms the other state-of-the-art methods on three benchmarks, offering improvements of up to 17.8$\%$ and 12.6$\%$ in terms of Edit distance and accuracy, respectively. Additionally, the NCA unit results in significantly better segmentation performance in the presence of only 5$\%$ labelled videos. We also demonstrate the generalizability and effectiveness of the proposed method on our Parkinson’s Disease Mouse Behaviour (PDMB) dataset. The code is publicly available at \href{https://github.com/FeixiangZhou/SMC-NCA}{https://github.com/FeixiangZhou/SMC-NCA}.
\end{abstract}


\section{Introduction}
\label{sec:intro}
\IEEEPARstart{A}{C}{T}{I}{O}{N}  recognition on trimmed videos has
achieved remarkable performance over the past few years \cite{yu2019weakly,9706285,9351626,9381645,9447897}. Despite the success of these approaches on trimmed videos with a single action, their ability to handle long videos containing multiple actions with different lengths is limited. Temporal action segmentation (TAS) aims at temporally detecting and recognising human action segments in a long untrimmed video \cite{lea2017temporal, shi2011human,li2020ms}, which has attracted a lot of attention in recent years. Different from temporal action detection (TAD) \cite{lin2019bmn} that detects the temporal boundaries (i.e., start and end) of action instances and predicts corresponding action categories simultaneously using temporally sparse action annotations, the goal of TAS is to produce frame-wise dense action labels. Besides, TAD operates with general videos from everyday life, such as THUMOS14 \cite{THUMOS14}, while TAS targets procedural activity datasets.
It is a key task for analysing and understanding human activities in complex long videos, and has been widely applied in fields such as video surveillance \cite{vishwakarma2013survey}, video summarization \cite{apostolidis2021video} and autonomous vehicles \cite{rasouli2019autonomous}.
The temporal relations among sequential human actions are crucial in TAS as they determine the sequence and timing of each action. Recent research \cite{zhang2024semantic2graph} reveals that semantic features based on label-text prompts can enhance action segmentation. Despite the impressive performance of the previous approaches \cite{farha2019ms,yi2021asformer,huang2020improving,zhang2024semantic2graph}, they rely on fully labelled videos where the start and end frames of each action are annotated, leading to substantial frame-wise annotation costs.
To alleviate this problem, many researchers have started exploring weakly supervised approaches using transcripts \cite{li2019weakly, lu2021weakly, souri2021fast}, action sets \cite{richard2018action,li2020set,fayyaz2020sct} or annotated timestamps \cite{li2021temporal} to reduce the annotation cost whilst maintaining action segmentation performance. However, significant efforts are needed to produce partial labels for supervised learning. Recently, semi-supervised approaches  \cite{singhania2022iterative, ding2022leveraging,singhania2023c2f} for this task have attracted increasing attention, with a small percentage of labelled videos in the training set.

\begin{figure}
\begin{center}
\includegraphics[width=9.3cm]{
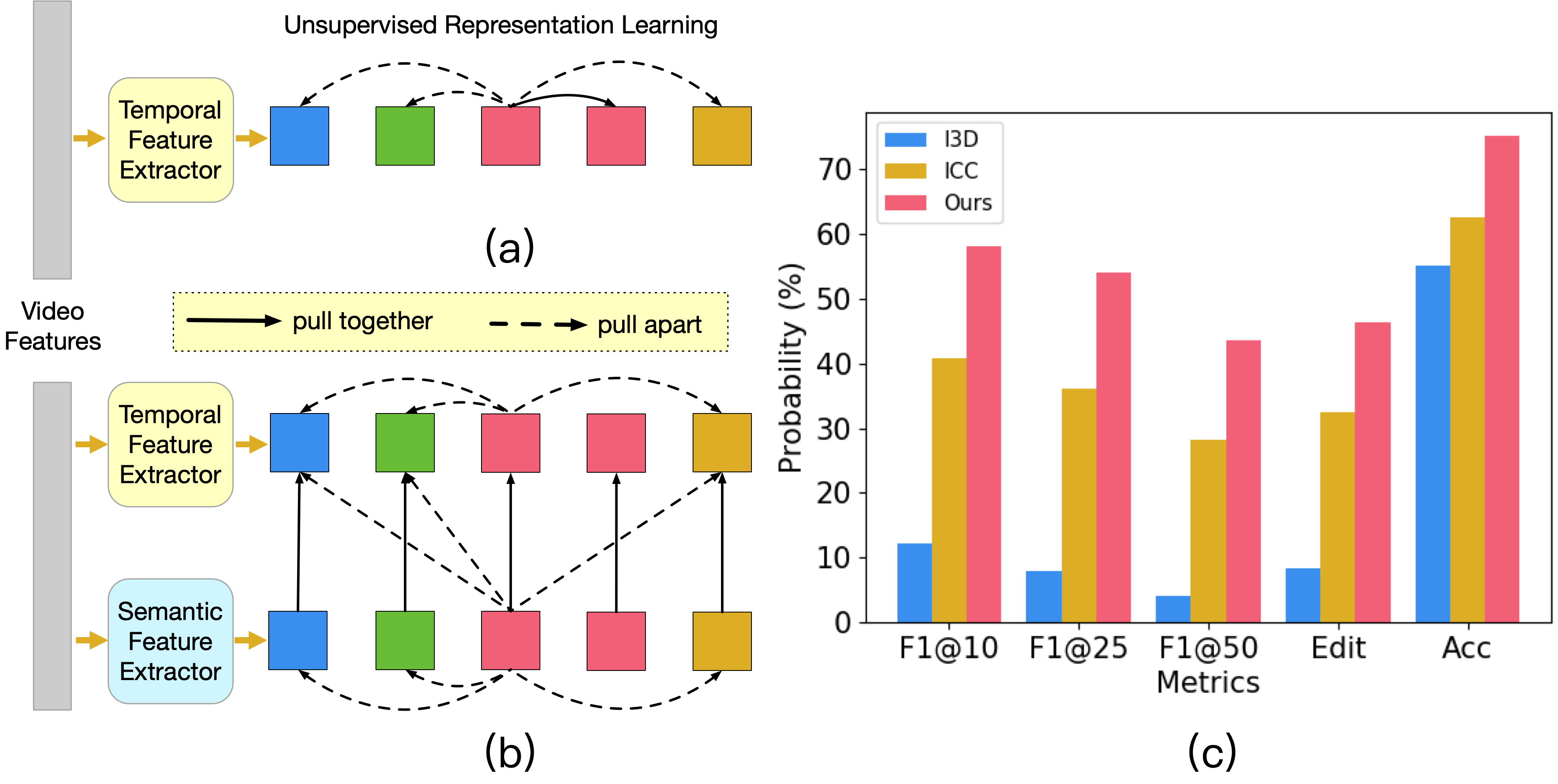}
\end{center}
\caption {Motivation of the proposed method. (a) ICC \cite{singhania2022iterative} only samples representations with temporal information for unsupervised contrastive representation learning, where the construction of both positive (red boxes with the same label) and negative  (boxes with different colours)  pairs is guided by input feature clustering. It is susceptible to clustering errors (see Fig. \ref{fig:evidence}), leading to weaker frame-wise representations.  
(b) Our method leverages both temporal and semantic information to fully explore intra- and inter-information variations. We form positive pairs by considering the inherent temporal-semantic consistency of each frame, as well as three types of negative pairs based on the dynamic clustering of the original input, temporal and semantic features, thereby mitigating the adverse effects of clustering errors.  (c) shows that our proposed method can learn more discriminative representations on the 50Salads dataset.}
\label{fig:comparison_ICC}
\end{figure}

\begin{figure}
\begin{center}
\includegraphics[width=8.3cm]{
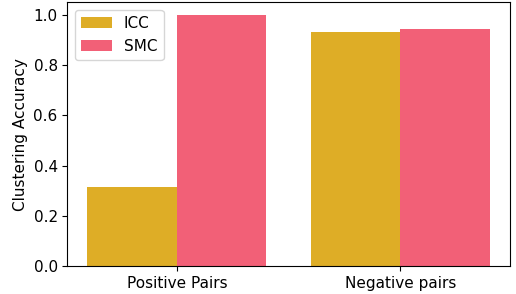}
\end{center}
\caption {Quantitative evidence that the quality of positive and negative pairs in ICC is affected by clustering errors. For both ICC and our SMC, we set the same batch size (5) and number of clusters (40), and the clustering accuracy is averaged over 50 epochs on the 50 Salads dataset by comparing the ground-truth labels with clustering labels. As observed, our SMC significantly improves the quality of positive pairs by leveraging the temporal-semantic consistency of each frame. The quality of negative pairs is also slightly enhanced by the dynamic clustering. }
\label{fig:evidence}
\end{figure}

SS-TAS is a non-trivial task as it often involves processing long untrimmed videos, presenting unique challenges compared to trimmed sequences. It can be difficult to effectively utilising the complex information contained within these unlabelled videos while ensuring accurate segmentation. ICC \cite{singhania2022iterative} is the first attempt to explore semi-supervised learning for TAS by introducing a contrast-classify framework, which consists of two steps, i.e., unsupervised contrastive representation learning for frame-wise representation enhancement, as well as semi-supervised learning for action classification. As shown in Fig. \ref{fig:comparison_ICC}(a), the contrastive learning approach in ICC primarily focuses on variations within temporal information containing relations between actions, which may ignore the potential role of action-specific semantic characteristics. Meanwhile, constructing both positive and negative sets for contrastive learning mainly depends on the clustering results (i.e., labels generated by clustering) of pre-trained input features. Empirical evidence has shown that the outcome of contrastive learning can be severely affected by clustering errors (see Fig. \ref{fig:evidence} and Tab. \ref{tab:ablation_unsuper} (the first set of experiments)). 
Another method for this topic is leveraging the correlation of actions between the labelled and unlabelled videos, and temporal continuity of actions \cite{ding2022leveraging}. However, their method only relies on action frequency prior in labelled data to build such correlations, rendering it inadequate for guiding the learning of complex unlabelled videos. Consequently, how to effectively leverage unlabelled, long untrimmed videos is still an open and challenging problem.

In this paper, we propose SMC-NCA, a novel temporal-semantic contrast framework for SS-TAS. Our work is also built on the contrast-classify architecture, but unlike ICC, which contrasts relevant frames within temporal information alone, our SMC incorporates both temporal information (i.e., temporal relations between actions) and semantic information (i.e.,  action-specific characteristics) into a unified contrastive learning framework, effectively reducing clustering errors and promoting frame-wise representation learning. We further design a NCA unit to handle continuously fluctuating action category predictions, a.k.a. over-segmentation issues, leading to better segmentation performance.

Inspired by the alignment of different augmented views of the same image/video \cite{chen2020simple,dorkenwald2022scvrl,zhu2022self}, as well as that of multiple modalities \cite{radford2021learning,zhang2024semantic2graph} (e.g., vision and language),  we incorporate a simple semantic feature extractor to extract semantic information, which complements the temporal information for cross-information contrastive representation learning. Semantic information emphasises action-specific characteristics \cite{zhang2024semantic2graph} in contrast to temporal information that encodes the dependencies between actions.  Hence, we argue that leveraging the two types of information can help to facilitate representation learning (see Fig.\ref{fig:comparison_ICC}(c)). As shown in Fig. \ref{fig:comparison_ICC}(b),
temporal and semantic branches are then jointly optimised by our SMC for fully exploring inter- and intra-information variations. Concretely, unlike \cite{singhania2022iterative} that recognises positive pairs based on the clustering results, SMC creates them (i.e., pairwise boxes connected by solid lines in Fig. \ref{fig:comparison_ICC}(b)) by directly aligning temporal-semantic information, relying on the inherent temporal-semantic consistency of each frame. Meanwhile, we explicitly construct three types of complementary negative pairs to enlarge inter-class dissimilarity, where the generation process is determined by our proposed dynamic clustering algorithm. With the proposed temporal-semantic contrast, our SMC pays attention to action-specific semantic features and temporal relations simultaneously, effectively reducing the impact of clustering errors, thus achieving enhanced representations (see Tab. \ref{tab:ablation_unsuper}).

Benefiting from strong representations learned by SMC, our approach offers a significant improvement on the semi-supervised frame-wise accuracy. However, there still exist significant over-segmentation errors  \cite{ishikawa2021alleviating} when we use a small amount of labelled data for training the system, resulting in low-quality segmentation results, i.e., segmental F1 score and Edit distance (see Tab. \ref{tab:ablatio_NCA}). This is because the existing contrast-classify framework excessively emphasises frame-wise classification performance by carefully performing inter-frame contrast to enhance the discriminability of each frame's representation, thus potentially overlooking the consistency of categories within a particular action segment.
To this end, we propose NCA to encourage the consistency between neighbourhoods (i.e., action segments) centered at different frames. Intuitively, feature distributions within the neighbourhoods of frames with the same label should be spatially close. NCA can be combined with SMC to deal with over-segmentation issues, particularly when we conduct supervised learning with a small amount of labelled data.

Our main contributions are highlighted as follows:
\begin{itemize}
\item We propose SMC-NCA, a novel temporal-semantic contrast framework tailored for SS-TAS, where temporal information with relations between actions, and semantic information with action-specific attributes are jointly utilised to enhance representation learning. 

\item We propose SMC for unsupervised representation learning, allowing one to fully explore intra- and inter-information variations for learning discriminative frame-wise representations with the support of semantic information.

\item We propose NCA to alleviate over-segmentation problems in semi-supervised settings by enforcing inter-neighbourhood consistency. This enables significant improvement in the segmentation quality while maintaining good frame-wise accuracy.

\item Experimental results demonstrate that our SMC-NCA outperforms state-of-the-art methods across different settings of labelled data (5$\%$, 10$\%$, 40$\%$ and 100$\%$) on three challenging datasets.

\item We also introduce a new Parkinson’s Disease Mouse Behaviour (PDMB) dataset, which provides a valuable resource for the research community to investigate the behavioural correlations of mice with Parkinson’s Disease. Experimental results show that our method is generalizable and effective on different kinds of datasets beyond human actions.

\end{itemize} 



\begin{figure*}
\begin{center}
\includegraphics[width=17cm]{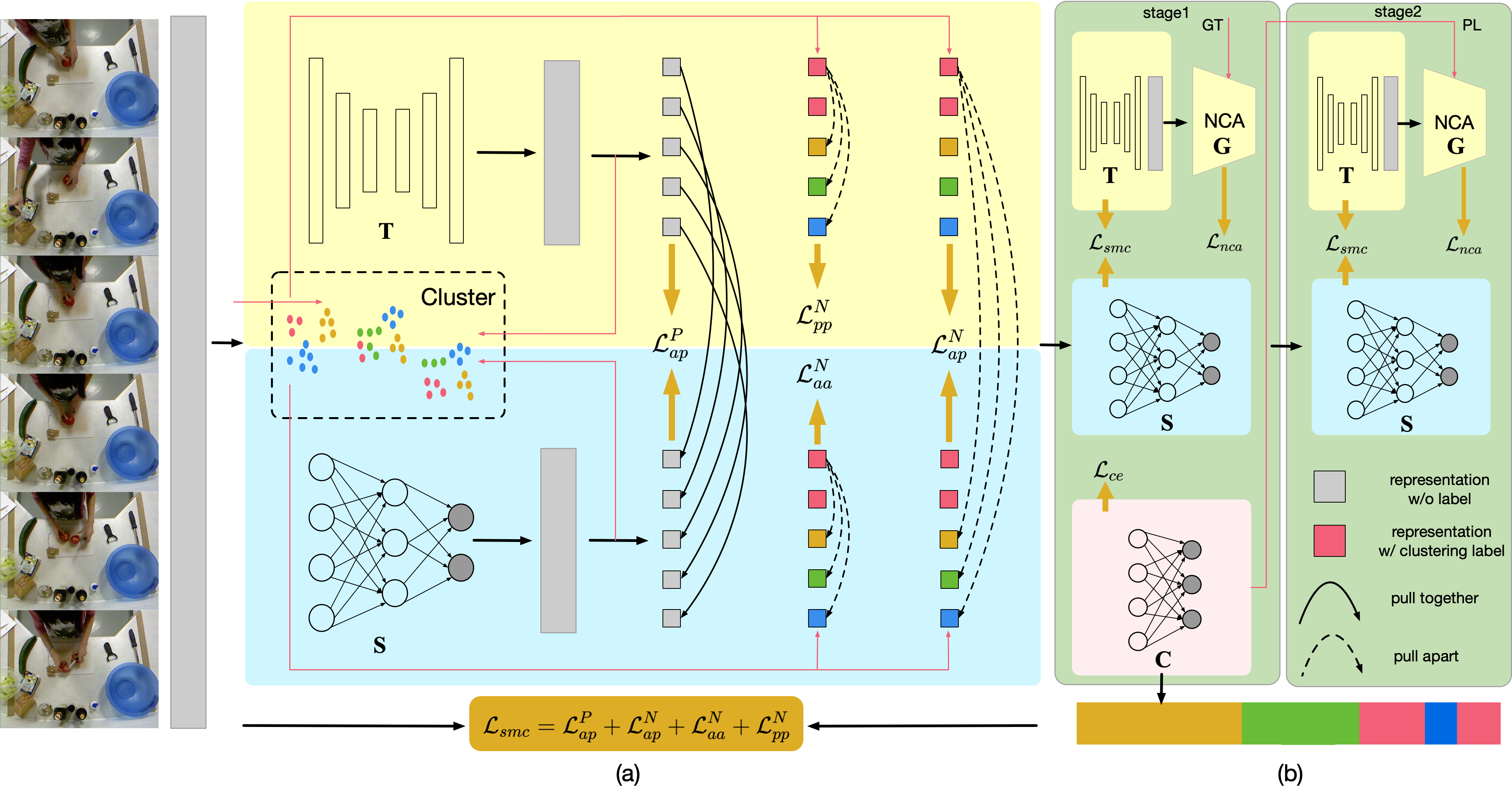}
\end{center}
\caption {Our proposed semi-supervised learning framework for temporal action segmentation. (a) Unsupervised representation learning. The pre-trained I3D features are fed into the long-term feature extractor $\textbf{T}$ and the semantic feature extractor $\textbf{S}$ to generate temporal and high-level semantic representations, respectively. Each kind of representation is sampled, followed by performing Semantic-guided Multi-level Contrast. (b) Semi-supervised learning. In stage 1, our Neighbourhood-Consistency-Aware unit $\textbf{G}$
works closely with SMC, and the classification layer $\textbf{C}$ can produce pseudo-labels (PL) used to guide the subsequent contrastive learning (stage 2). $\mathcal{L}_{ap}^{N}$, $\mathcal{L}_{aa}^{N}$ and $\mathcal{L}_{pp}^{N}$ denote the negative losses of three negative pairs which are driven by the clustering results. $\mathcal{L}_{ap}^{P}$ represents the positive loss without relying on the clustering labels. $\mathcal{L}_{nca}$ is generated in the NCA module. Note that T and
C are two sub-networks of the C2F-TCN \cite{singhania2021coarse,singhania2022iterative} backbone. The former is responsible for temporal feature extraction, while the latter is connected after the former to perform action classification. $\textbf{S}$ and $\textbf{G}$ are only used for training in order to facilitate learning and will not be used during the inference stage.}
\label{fig:framework}
\end{figure*}

\section{Related Work}
\label{sec:relatedwork}

\subsection{Temporal Action Segmentation}
\label{sec:section2.1}
Temporal action segmentation (TAS) has been an increasingly popular trend in the domain of video understanding, which involves different levels of supervision. Fully-supervised methods require that each frame in the training videos be labelled.  Earlier approaches have attempted to incorporate high-level temporal modelling over frame-wise classifiers. Kuehne et al. \cite{kuehne2016end} utilised Fisher vectors of improved dense trajectories to represent video frames, and modelled each action with a Hidden Markov Model (HMM). Singh et al. \cite{singh2016multi} used a two-stream network to learn representations of short video chunks, which are then fed into a bi-directional LSTM to capture dependencies between different chunks. However, their approaches are computationally intensive due to sequential prediction. Most recent approaches employ TCN (temporal convolutional network)  \cite{lea2017temporal,farha2019ms,singhania2021coarse}, GCN (graph convolution network)  \cite{huang2020improving, wang2021temporal} or Transformer  \cite{yi2021asformer} to model the long-range temporal dependencies in the videos. In spite of their promising performance, obtaining annotations on fine-grained actions for each frame is costly. Weakly-supervised methods mainly focus on transcript-level (with action order information)  \cite{li2019weakly, lu2021weakly,souri2021fast}, set-level (without action order information)  \cite{richard2018action,li2020set,fayyaz2020sct} and timestamp-level  \cite{li2021temporal} supervisions to reduce the annotation effort. However, necessary supervision information is also needed for each video in the training set. Unsupervised approaches  \cite{kukleva2019unsupervised,sarfraz2021temporally, kumar2022unsupervised,ding2022temporal} take advantage of clustering algorithms without any supervised signal per video, which still suffer from poor performance compared to other supervised settings.

\subsection{Unsupervised Contrastive Representation Learning}
\label{sec:section2.2}
Visual representation plays a crucial role in various computer vision tasks, including temporal action segmentation \cite{singh2016multi,yi2021asformer}, person retrieval \cite{shi2020adaptive,shi2020person} and event detection \cite{shen2008modality}.
It can be learned in different ways, including fully supervised, semi-supervised and unsupervised methods. For instance, Shi et al. \cite{shi2020person} proposed to mine various human attributes and model relations between them to learn precise attribute representations. Shen et al. \cite{shen2008modality} explored multi-modal information fusion to enrich video content representations.

Recently, contrastive representation learning has been extensively used in different computer vision applications such as image representation learning \cite{he2020momentum, chen2020simple,oord2018representation,tian2020contrastive}, video representation learning \cite{dorkenwald2022scvrl,hu2021contrast,qian2021spatiotemporal,recasens2021broaden,biswas2021multiple, zhu2022self}, time series representation learning \cite{eldele2021time,yue2022ts2vec,franceschi2019unsupervised}, face recognition \cite{schroff2015facenet,ge2018deep,boutros2022self}. Among these approaches, contrastive loss\cite{chen2020simple,he2020momentum} and triplet loss \cite{schroff2015facenet,franceschi2019unsupervised} are two popular loss functions to perform contrasting. The underlying idea behind the former is to pull together representations of augmented samples of the same image or video clips (i.e., positive pair) while pulling apart those of different instances (i.e., negative pair). The latter has the same goal but involves defining a triplet of (\textit{anchor, positive and negative pairs}), where the positive pairs (i.e., anchor and positive representations) should be close and the negative pairs (i.e., anchor and negative representations) should be far apart.

The idea of visual-semantic alignment for representation enhancement has been explored in various fields such as few-shot learning \cite{zhang2023deta,afham2022visual}, meta learning \cite{zhang2022progressive} and out-of-distribution detection \cite{zhang2023global}. Different from most methods that operate within a supervised setting where label information is utilised to guide feature/image alignment, our SMC performs temporal-semantic contrast in a unsupervised way to fully utilise unlabelled videos. Furthermore, it is based on the triplet loss \cite{franceschi2019unsupervised}, which effectively reduces the dependence on noisy clustering labels, and leverages complementary information between temporal and semantic information compared with other methods.

\subsection{Semi-supervised Learning}
Semi-supervised learning (SSL) is a good solution to reduce the cost of data labelling, which leverages a few labelled samples and a large number of unlabeled samples to train the model. The current semi-supervised methods are mainly divided into prediction-based and representation-based approaches \cite{mo2023ropaws}. Prediction-based approaches typically involve regularizing the classifier's prediction of unlabelled data, where two common paradigms for this purpose are pseudo-labelling \cite{lee2013pseudo,xie2020self,rizve2021defense} and consistency regularization \cite{sohn2020fixmatch,miyato2018virtual,tang2021humble}. Pseudo-labelling based methods generate pseudo labels for unlabeled data through pre-trained models and use these pseudo labels to further optimise the model. Consistency based methods encourage the model to make similar predictions on images obtained by applying different data augmentation techniques to the same image.
For example, FixMatch \cite{sohn2020fixmatch} uses the model to generate pseudo labels for weakly augmented unlabeled images and only samples with highly-confident pseudo-labels will be used for training. To adapt FixMatch to video-based action recognition, Wu et al.  \cite{wu2023neighbor} designed a neighbor-guided consistent and contrastive learning framework to reuse lowly-confident samples and learn more discriminative feature representations.
However,  prediction-based approaches are less effective compared to representation-based approaches in large-scale scenarios \cite{mo2023ropaws}.

Representation-based methods focus more on the representations encoded by deep learning networks. 
Recent studies exploit the success of self-supervised learning \cite{he2020momentum, chen2020simple} in learning representations from unlabeled data to train large-scale semi-supervised models \cite{chen2020big,cai2022semi}. PAWS \cite{assran2021semi} proposed a single-stage training scheme that combines supervised and self-supervised learning for SSL. Although these methods have shown promise, it remains challenging to directly apply data augmentation strategies, such as flipping, rotation, and transformation, used in the image or video domain to TAS as the inputs are pre-computed feature vectors \cite{ding2022leveraging}. 
Inspired by \cite{singhania2022iterative}, we propose a novel temporal-semantic contrast framework for representation enhancement, which shows strong frame-wise representations and segmentation performance.


\section{Proposed Method}
\label{sec:section3}
The overall framework of our SMC-NCA is illustrated in Fig. \ref{fig:framework}. For unsupervised representation learning, given the input video features, a temporal feature extractor $\textbf{T}$ and semantic feature extractor $\textbf{S}$ are jointly adopted to extract temporal and semantic information. Our SMC learns not only intra-information variation but also inter-information variation, where the temporal information describes the temporal dependencies between actions while semantic information highlights the action-specific characteristics. For semi-supervised learning, the models $\textbf{T}$, $\textbf{S}$ and a linear classifier $\textbf{C}$ are firstly integrated to perform supervised learning on a small number of labelled videos, followed by a new round of contrastive learning based on the pseudo-labels provided by $\textbf{C}$. An additional NCA unit ($\textbf{G}$) is designed in conjunction with contrastive learning to measure the spatial consistency between different neighbourhoods for the alleviation of over-segmentation errors. Following \cite{singhania2022iterative}, we 
 iteratively perform classification and contrastive learning.  
 We have not altered the structure of the backbone C2F-TCN \cite{singhania2022iterative}; instead, we introduce two additional modules, $\textbf{S}$ and $\textbf{G}$, to facilitate unsupervised and semi-supervised learning. These two modules seamlessly integrate with the backbone and are exclusively utilised during model training to improve the semi-supervised segmentation performance. They are not utilised during the inference phase.

\subsection{Preliminaries on Contrastive Learning}

\textbf{Contrastive loss}. InfoNCE loss \cite{oord2018representation,chen2020simple,he2020momentum} is usually used for optimisation in unsupervised representation learning. It is calculated on images or video clips with the common goal of encouraging instances of the same class to be approaching, and pushing the instances of different classes apart from each other in the embedding space. Following the formalism of \cite{chen2020simple, singhania2022iterative}, we define a set of features  $\mathcal{F}=\left\{\mathbf{f}_{i}| i \in \mathcal{I}\right\}$, which is represented as a matrix $\mathbf{F}$ ($\mathbf{F}[i]=\mathbf{f}_{i}
$). $\mathcal{I}$ denotes the set of feature indices and each feature $\mathbf{f}_{i}$ has an unique class. Thus, given $\mathbf{F}[i]$, the positive set $\mathcal{P}_{i} \subset \mathcal{I}$ consisting of indices of features that have the same class as $\mathbf{F}[i]$, and the corresponding negative set $\mathcal{N}_{i}$, the contrastive loss for each $j\in\mathcal{P}_{i}$ is defined as:
\begin{equation}
\begin{split}
\mathcal{L}_{cont}(i,j) 
=-\log \frac{e^{\operatorname{sim}\left(\mathbf{F}[i],\mathbf{F}[j]\right) / \tau}}{e^{\operatorname{sim}\left(\mathbf{F}[i],\mathbf{F}[j]\right) / \tau}+\sum\limits_{k\in\mathcal{N}_{i}}  e^{\operatorname{sim}\left(\mathbf{F}[i],\mathbf{F}[k]\right) / \tau}}
\label{equation:con_loss}
\end{split}
\end{equation}
where $\operatorname{sim}(\mathbf{F}[i], \mathbf{F}[j])=\mathbf{F}[i]^{\top} \mathbf{F}[j] /\|\mathbf{F}[i]\|_{2}\|\mathbf{F}[j]\|_{2}$ is the inner product between two $\ell_{2}$ normalised vectors. $\tau>0$ is a temperature parameter. 

\textbf{Triplet loss}. Unlike the contrastive loss, triplet loss \cite{schroff2015facenet, franceschi2019unsupervised} requires to define a triplet $(\mathbf{F}[a],\mathbf{F}[p],\mathbf{F}[n] )
$ where $\mathbf{F}[a]$, $\mathbf{F}[p]$ and $\mathbf{F}[n]$ represent the anchor, positive and negative representations, respectively. This loss attracts positive pairs, $\mathbf{F}[a]$ (anchor) and $\mathbf{F}[p]$ (positive), while pushing away negative pairs, $\mathbf{F}[a]$ and $\mathbf{F}[n]$ (negative). Following \cite{franceschi2019unsupervised}, the triplet loss is formulated as:
\begin{equation}
\begin{split}
\mathcal{L}_{trip}(a,p,n) =&-\log\Big(\sigma(\mathbf{F}[a]^{\top }\mathbf{F}[p] )\Big)\\ &-\log\Big(\sigma(-\mathbf{F}[a]^{\top }\mathbf{F}[n] )\Big) 
\label{equation:trip_loss}
\end{split}
\end{equation}
where $\sigma$ is the sigmoid function. The core of the triplet loss is to effectively generate the three representations.

\subsection{Problem Definition}
A video sequence can be represented as $\textbf{V}\in \mathbb{R}^{T\times E} $ where $T$ is the total number of frames and $E$ represents the dimension of the pre-trained I3D features \cite{carreira2017quo}. Each video frame $\textbf{V}[t]\in\mathbb{R}^{E}$ has a ground-truth action label $y[t] \in\{1, \ldots, A\} $ where $A$ denotes the number of the classes. We use an encoder-decoder model, i.e.,  C2F-TCN \cite{singhania2021coarse} as our base segmentation network $\textbf{T}$. Therefore, given the input video features $\textbf{V}$, we aim to learn the unsupervised model $\left ( \textbf{T}: \textbf{S}\right ) $ by our Semantic-guided Multi-level Contrast on the dataset $\mathcal{D}=\mathcal{D}_{U} \cup \mathcal{D}_{L}$ ($\mathcal{D}_{U}$ and $\mathcal{D}_{L}$ denote the unlabelled and labelled videos, respectively) where $\textbf{T}$ and $\textbf{S}$ are the temporal and semantic models, respectively. Subsequently, the  Neighbourhood-Consistency-Aware module $\textbf{G}$ and a linear mapping layer $\textbf{C}$ are added to form semi-supervised model $\left ( \textbf{T}: \textbf{S}:\textbf{G}:\textbf{C}\right ) $ which is trained on a small subset of labelled training videos $\mathcal{D}_{L}$. We follow the standard operation of \cite{singhania2022iterative}, where we first use the same down-sampling strategy to obtain coarser features (i.e., input) of temporal dimension $T$. In the inference, the final predictions are up-sampled to the original full resolution with temporal dimension $T_{ori}$ to compare with the original ground truth. 

\subsection{Semantic-guided Multi-level Contrast}
\label{sec:section3.3}
As mentioned earlier, we propose to leverage both temporal and semantic information to our proposed SMC scheme, where inter- and intra-information variations are jointly learned to enhance frame-wise representations. 

\textbf{Temporal and semantic information generation}. To encode the correlation between actions, we adopt the C2F-TCN model to produce the multi-resolution representation and combine representations of different resolutions into a new representation using the same temporal up-sampling approach in \cite{singhania2022iterative}. Similar to \cite{singhania2022iterative}, the combined representation with temporal information is used for unsupervised contrastive learning, which is denoted as $\textbf{X}=\textbf{T}(\textbf{V})\in \mathbb{R}^{N\times T\times D}$, where $N$ is the mini-batch size during training and $D$ represents the feature dimension. 

In ICC, only temporal information is utilised to perform inter-frame contrastive learning. In contrast, we propose to incorporate semantic information alongside temporal information for cross-information contrast. This decision is motivated by the significance of semantic features in capturing action-specific details, a factor that has been proven to be effective in promoting temporal action segmentation \cite{zhang2024semantic2graph}. Different from extracting semantic features from action- or class-specific text information \cite{zhang2024semantic2graph}, we implicitly extract such features directly from the raw input features without the need for complex text encoding models. 
Specifically, we employ the multilayer perceptron (MLP) as a semantic feature extractor on the input video features $\textbf{V}$ to generate a new representation (denoted as  $\textbf{H}=\textbf{S}(\textbf{V})\in \mathbb{R}^{N\times T\times D}$) with high-level semantic information. A MLP offers simplicity and efficiency (we also test other extractors, such as deeper MLPs and 1D CNNs, see Tab. \ref{tab:comparison_mlp}).  The model $\textbf{S}$ can be formulated as: 
\begin{equation} 
\begin{split}
\mathbf{V}^{(l+1)}&=\sigma\left(\mathbf{V}^{(l)} \mathbf{W}^{(l)}+b^{(l)}\right) \\ 
\mathbf{H}&=\mathbf{V}^{(l+1)} \mathbf{W}^{(l+1)}+b^{(l+1)}
\label{equation:mlp}
\end{split}
\end{equation}
where $\mathbf{V}^{(l)}\in \mathbb{R}^{N\times T\times D^{l}} $ denotes the input hidden representation from the previous layer (with $\mathbf{V}^{(0)}=\mathbf{V}$) and $\mathbf{W}^{(l)}\in \mathbb{R}^{D^{l}\times D^{l+1}}$ and $b^{(l)}$ are trainable parameters. $\sigma$ is an activation function. 
Temporal and semantic information complement each other in promoting representation learning as they capture different aspects of actions. Next, we will elaborate on how to incorporate these two types of information into a unified contrastive learning framework.

\textbf{Inter-information variation learning}. 
Under the guidance of clustering labels, ICC reduces the intra-class distance by bringing frames belonging to the same label closer together. However, it is susceptible to clustering errors. In practice, the distribution of data may be complex and uneven, leading the clustering algorithm to assign similar frames to different clusters or dissimilar frames to the same cluster. This clustering error can cause  ICC to learn incorrect similarity information during training, thus affecting its performance. Additionally, clustering errors may result in overfitting to samples with noisy labels, thereby reducing its generalization ability.

With the support of semantic information, we directly construct inter-information positive pairs without relying on noisy clustering labels. To be more specific, we firstly sample a fixed number of frames from each training video to form a new temporal representation $\textbf{X}_{s} \in \mathbb{R}^{N\times T_{s}\times D}$ and semantic representation $\textbf{H}_{s} \in \mathbb{R}^{N\times T_{s}\times D}$ using the same sampling strategy reported in \cite{singhania2022iterative}, where $T_{s}$ is the number of the sampled frames from each video. This is because it is too computationally expensive to measure each frame of every video in a mini-batch. The sampled frames for each video are then combined by concatenation operation on the dimension $T$ to form $\textbf{X}_{s} \in \mathbb{R}^{N\cdot T_{s}\times D}$  and $\textbf{H}_{s} \in \mathbb{R}^{N\cdot T_{s}\times D}$.   

The features of pairwise video frames with the same index in $\textbf{X}_{s}$ and $\textbf{H}_{s}$ describe the same action (boxes connected by solid lines in Fig. \ref{fig:comparison_ICC}(b)). In this way, we identify $\textbf{X}_{s}$ and $\textbf{H}_{s}$ as positive and anchor representations respectively, and our goal is to pull them (i.e., positive pairs) together. Inspired by \cite{franceschi2019unsupervised}, given  $\textbf{X}_{s}$ and $\textbf{H}_{s}$, we formalise the objective function of the inter-information similarity learning as:
\begin{equation} 
\begin{split}
\Psi_{ap}^{P} &=-\Big[\log\Big(\sigma(\mathbf{H}_{s}^{\top }\mathbf{X}_{s}/\xi)\Big)\Big]\odot \mathbf{I}
\\&=\begin{bmatrix}
  \Psi_{ap}^{P}[1,1]   &  & \\
  & \ddots   & \\
  &  & \Psi_{ap}^{P}[N\cdot T_{s},N\cdot T_{s}]  \\
\end{bmatrix}
\label{equation:P_ap}
\end{split}
\end{equation}
where $\odot$ represents the element-wise product. $\mathbf{I}\in \mathbb{R}^{N\cdot T_{s}\times N\cdot T_{s}}$ represents the identity matrix used to select pairwise frames with the same index, and $0<\xi<1$ is a scale factor to adjust the correlation between vectors. $ \Psi_{ap}^{P}[,] $ is the element of matrix  $\Psi_{ap}^{P}$. 

The advantages of optimising this function are twofold. Firstly, we consider the temporal and semantic information simultaneously, which is beneficial to improving the ability of representation learning. Secondly, learning of positive pairs (anchor and positive representations) is not dependent on the coarse clustering results, compared to ICC which generates related contrastive instances based on clustering results. In contrast, $\Psi_{ap}^{P}$ performs contrastive learning with the guide of the potential consistency between frame-wise temporal and semantic representations. We then compute the inter-information similarity loss $\mathcal{L}_{ap}^{P}$ between positive pairs below:
\begin{equation} 
\begin{split}
\mathcal{L}_{ap}^{P}=\frac{1}{N\cdot T_{s}}\sum_{i=1}^{N\cdot T_{s}}(\Psi_{ap}^{P}[i,i])  
\label{equation:P_ap_loss}
\end{split}
\end{equation}
where $\mathcal{L}_{ap}^{P}$ promotes intra-class similarity by considering the one-to-one correspondence between temporal and semantic information, where each frame located at the same position belongs to the same category.

Operations such as random selection \cite{franceschi2019unsupervised} or shuffling \cite{dorkenwald2022scvrl} are usually used to generate negative pairs in the existing works for unsupervised contrastive learning \cite{ming2017simple,biswas2021multiple,franceschi2019unsupervised,boutros2022self}. However, these strategies are usually used for short temporal sequences and they cannot be directly applied to temporal action segmentation due to the complex and various actions in long videos. Here, we further exploit two types of complementary information i.e., temporal and semantic representations to create negative pairs for uncovering their differences. Different from learning inter-information similarity without any auxiliary label information shown in Eq. (\ref{equation:P_ap}), the construction of negative pairs relies on the results of K-means clustering. Specifically, we can choose pairwise frames with different clustering labels, followed by contrasting these frames. For the $t$-frame in  $\textbf{V}_{s}$, the clustering label generated by K-means can be denoted as $l_{in}[t]$. We can further obtain a matrix, i.e., $\textbf{M}_{in}$, which consists of the similarity between any two frames. Regarding the matrix $\textbf{M}_{in}$, we have: 
\begin{equation}  
\begin{split}
\textbf{M}_{in}[i,j]=\begin{cases}
  1 \text{, } l_{in}[i]=l_{in}[j],\forall i,j\in\left \{ 1,2,\dots ,N\cdot T_{s} \right \}    \\
  0 \text{, } other
\end{cases}
\label{equation:matrx}
\end{split}
\end{equation}  
where $\textbf{M}_{in}$ indicates the similarity between all sampled frames in a mini-batch. When the clustering labels between the $i$-th and $j$-th frames are identical, the corresponding element in the matrix is set to 1; otherwise, it is set to 0.

The clustering results on the standard input features $\textbf{V}_{s}$ cannot be updated during training \cite{singhania2022iterative}, which may lead to accumulation errors caused by incorrect negative pairs. To reduce these errors, we perform the clustering on the sampled input $\textbf{V}_{s}$, temporal $\textbf{X}_{s}$ and semantic $\textbf{H}_{s}$ features simultaneously for dynamic selection of potential negative pairs. 
Hence, we obtain $\textbf{M}_{te}$ and $\textbf{M}_{se}$ for $\textbf{X}_{s}$ and $\textbf{H}_{s}$, respectively. By combining the three matrices, we generate a matrix $\textbf{M} \in \mathbb{R}^{N\cdot T_{s}\times N\cdot T_{s}}$ that is used to dynamically guide the dissimilarity learning on the negative pairs, defined as:    
\begin{equation} 
\begin{split}
\textbf{M}=(1-\textbf{M}_{in})\odot(1-\textbf{M}_{te})\odot (1-\textbf{M}_{se})
\label{equation:kmeans}
\end{split}
\end{equation}  
where each element $\textbf{M}[i,j]\in\left \{ 0,1 \right \} $  represents whether the $i$-th and $j$-th frames share the same clustering category. 1 indicates frames from different categories while 0 indicates frames from the same category. The combination of these three matrices can help improve the quality of negative pairs.

Afterwards, we can construct a set of dense negative pairs for each frame based on $\textbf{M}$, and negative pairs must be separate. Hence, similar to Eq. (\ref{equation:P_ap}), the function $\Psi_{ap}^{N}$ to be minimised with the corresponding inter-information dissimilarity loss $\mathcal{L}_{ap}^{N}$ is expressed as:
\begin{equation} 
\begin{split}
\Psi_{ap}^{N} &=-\Big[\log\Big(\sigma(-\mathbf{H}_{s}^{\top }\mathbf{X}_{s}/\xi   )\Big)\Big]\odot \mathbf{M}
\label{equation:N_ap}
\end{split}
\end{equation}
\begin{equation} 
\begin{split}
\mathcal{L}_{ap}^{N}=\frac{1}{N_{ap}}\sum_{i=1}^{N\cdot T_{s}}\sum_{j=1}^{N\cdot T_{s}}(\Psi_{ap}^{N}[i,j])  
\label{equation:N_ap_loss}
\end{split}
\end{equation}
where $\mathbf{M}$ is utilised to select pairwise frames between temporal and semantic information to construct inter-information negative pairs. $N_{ap}=\sum_{i=1}^{N\cdot T_{s}}\sum_{j=1}^{N\cdot T_{s}}(\mathbf{M}[i,j])$ denotes the number of all the negative pairs. $\mathcal{L}_{ap}^{N}$ facilitates the separation of inter-information different categories.

 \textbf{Intra-information variation learning}. $\mathcal{L}_{ap}^{P}$ and $\mathcal{L}_{ap}^{N}$ defined in Eqs.  (\ref{equation:P_ap_loss}) and (\ref{equation:N_ap_loss}) encourage the temporal model to focus on inter-information variations (i.e., similarity and dissimilarity). Although they implicitly learn the dissimilarity between frames with the semantic or temporal representation, the discriminability of the learned representations is still weak as the inherent discriminative cues within semantic or temporal information are not exploited (this can be verified in Section \ref{sec:section4.3.1}). 
 
 Therefore, we further explore the intra-information variations, including the dissimilarities within the semantic or temporal representation. In detail, for each frame of semantic representation (anchor), we choose all the other frames with different clustering labels to form a set of dense intra-information negative pairs. The intra-information dissimilarity loss based on semantic representation $\mathbf{H}_{s}$ can be computed as:
\begin{equation} 
\begin{split}
\Psi_{aa}^{N} &=-\Big[\log\Big(\sigma(-\mathbf{H}_{s}^{\top }\mathbf{H}_{s}/\xi   )\Big)\Big]\odot \mathbf{M}
\label{equation:N_aa}
\end{split}
\end{equation}
\begin{equation} 
\begin{split}
\mathcal{L}_{aa}^{N}=\frac{1}{N_{ap}}\sum_{i=1}^{N\cdot T_{s}}\sum_{j=1}^{N\cdot T_{s}}(\Psi_{aa}^{N}[i,j])  
\label{equation:N_aa_loss}
\end{split}
\end{equation}
where $\sigma$ is the sigmoid function.   $\mathbf{M}$, $\xi$ and $N_{ap}$ are defined in Eqs. (\ref{equation:kmeans}), (\ref{equation:P_ap}) and (\ref{equation:N_ap_loss}), respectively. Here $\mathbf{M}$ is used to select pairwise frames with distinct categories from semantic representation $\mathbf{H}_{s}$. $\mathcal{L}{aa}^{N}$ encourages the separation of different categories within semantic information. Additionally, we consider the dissimilarity between frames of temporal representation $\mathbf{X}_{s}$ (positive), and the corresponding loss is defined as follows:     
\begin{equation} 
\begin{split}
\Psi_{pp}^{N} &=-\Big[\log\Big(\sigma(-\mathbf{X}_{s}^{\top }\mathbf{X}_{s}/\xi   )\Big)\Big]\odot \mathbf{M}
\label{equation:N_pp}
\end{split}
\end{equation} 
\begin{equation}
\begin{split}
\mathcal{L}_{pp}^{N}=\frac{1}{N_{ap}}\sum_{i=1}^{N\cdot T_{s}}\sum_{j=1}^{N\cdot T_{s}}(\Psi_{pp}^{N}[i,j])  
\label{equation:N_pp_loss}
\end{split}
\end{equation}

By integrating inter-information losses (i.e., Eqs. (\ref{equation:P_ap_loss}) and  (\ref{equation:N_ap_loss})) with the intra-information losses in Eqs. (\ref{equation:N_aa_loss}) and (\ref{equation:N_pp_loss}), our proposed sematic-guided multi-level contrast loss is formulated as:
\begin{equation} 
\begin{split}
\mathcal{L}_{smc}=\mathcal{L}_{ap}^{P}+\mathcal{L}_{ap}^{N}+\mathcal{L}_{aa}^{N}+\mathcal{L}_{pp}^{N}
\label{equation:cmls_loss}
\end{split}
\end{equation}


Notably, by introducing semantic information to the temporal-semantic contrast framework, our method mines more useful information in a large number of unlabelled untrimmed videos. Semantic information plays a multifaceted role in our method: (1) It contributes to the construction of inter-information positive pairs by aligning temporal and semantic information directly, thereby enlarging the intra-class similarity between the two types of information. (2) It also participates in constructing inter-information negative pairs, aiming to enlarge inter-class dissimilarity between the two types of information. (3) It is involved in constructing intra-information negative pairs, further enlarging the intra-class dissimilarity within semantic information. In summary, with the support of semantic information, our SMC can learn representations that capture both the temporal dynamics and semantic nuances of actions, thereby improving the discriminability and robustness of these representations.

\begin{figure}
\begin{center}
\includegraphics[width=7.3cm]{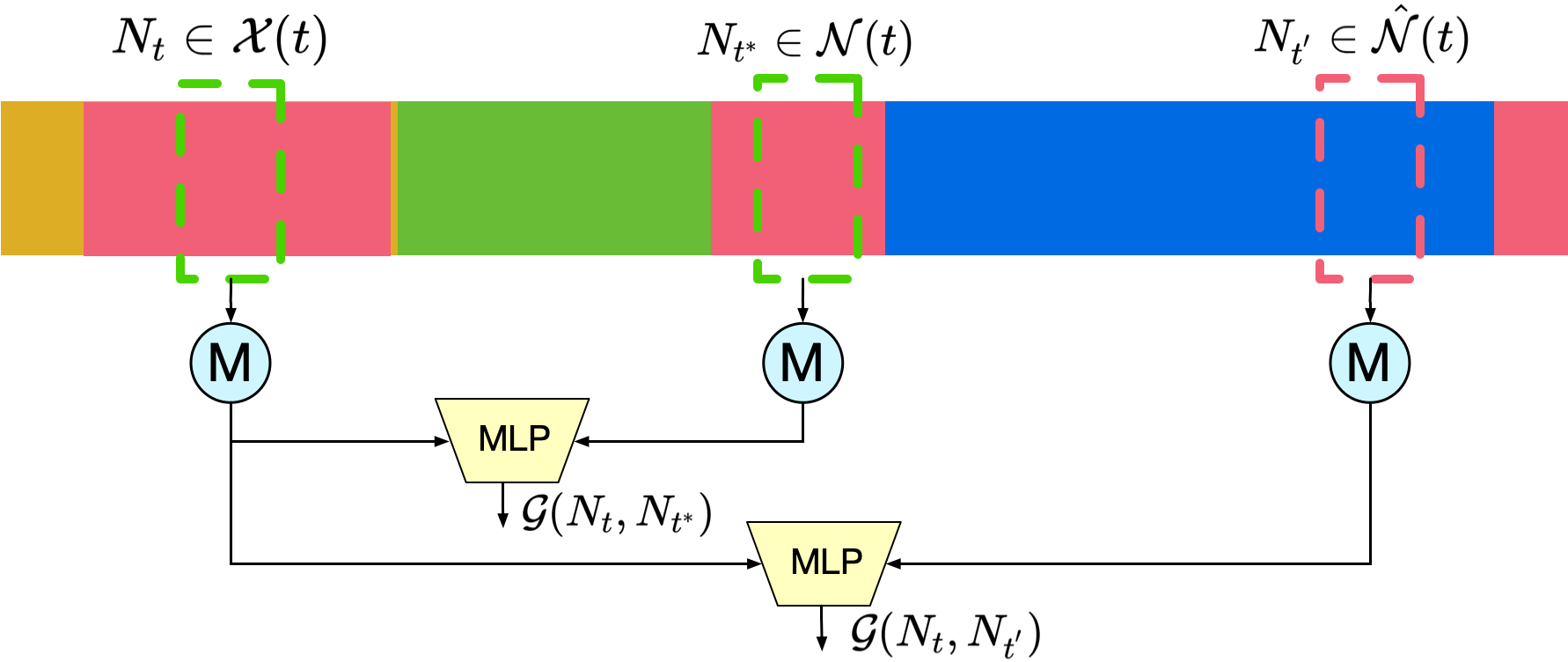}
\end{center}
\caption[Illustration of the proposed Neighbourhood-Consistency-Aware Unit] {Illustration of the proposed Neighbourhood-Consistency-Aware Unit. We randomly sample a frame $t$, and its neighbourhood is $N_{t}$ (left green dashed box). Then we can use frame $t^{*}$ with the same label to form the neighbourhood $N_{t^{*}}$. The two neighbourhoods are fed into the max-pooling layer to encode their feature distributions, followed by predicting the probability of having similar feature distributions by the MLP model. The same operation is applied to $N_{t}$ and $N_{t^{'}}$ centered at frames $t$ and $t^{'}$ with different labels.       }
\label{fig:NCA}
\end{figure}

\subsection{Neighbourhood-Consistency-Aware Unit}
\label{sec:section3.4}
Similar to \cite{singhania2022iterative}, after having trained the unsupervised model $\left ( \textbf{T}: \textbf{S}\right ) $ by our SMC, we perform semi-supervised learning for frame-wise classification of temporal actions using few labelled data, as shown in Fig. \ref{fig:framework}(b). Although this method enjoys good frame-wise accuracy, it suffers from significant over-segmentation errors, leading to low segmentation quality, as measured by segmental F1 score and Edit distance. The reason for this discrepancy lies in the fact that the pre-trained model focuses more on frame-level classification and does not consider the consistency of class labels within an action segment.
Over-segmentation refers to the error type where a video is excessively divided into subsegments, causing erratic action class predictions. To alleviate this issue in semi-supervised settings, we introduce a Neighbourhood-Consistency-Aware (NCA) module shown in Fig. \ref{fig:NCA} to ensure spatial consistency between action segments. The intuition behind this is that feature distribution within the neighbourhoods of the same label should be spatially close.


Concretely, at the first stage shown in Fig. \ref{fig:framework}(b), given the temporal representation  $\textbf{X}$ of a video with ground-truth labels $l$ in the batch, we randomly sample $K$ frames and construct $K$ neighbourhoods centered at these frames. For each frame, we then select $M$ frames having the same label and $M$ frames with different labels, respectively. Formally, we define the sets of neighbourhoods as:            
\begin{equation} 
\begin{split}
\mathcal{X}(t)&=\left\{N_{t}=\mathbf{X}[t-\frac{W}{2},t+\frac{W}{2} ]| t \in \mathcal{Z}(t)\right\}
\\
\mathcal{N}(t)&=\left\{N_{t^{*}}=\mathbf{X}[t^{*} -\frac{W}{2},t^{*}+\frac{W}{2} ]| t^{*} \in \mathcal{P}(t)\right\}
\\
\hat{\mathcal{N}}(t)&=\left\{N_{t^{'}}=\mathbf{X}[t^{'} -\frac{W}{2},t^{'}+\frac{W}{2} ]| t^{'} \in \hat{\mathcal{P}}(t)\right\}
\label{equation:NCA_sets}
\end{split}
\end{equation}
where $W$ is the length of an action segment. $\mathcal{N}(t)$ represents a set of $M$ segments, where each segment consisting of $W$ frames is centered at one frame from a group of $M$ frames. The $M$ frames are randomly sampled from the video frames that have the same label as the frame $t$. For $\hat{\mathcal{N}}(t)$, $M$ frames are randomly sampled from the video frames that have different labels from frame $t$. $\mathcal{Z}(t)=\left \{ t_{i}|1\le i \le K\right \}$ represents the indices of the sampled $K$ frames. $\mathcal{P}(t)=\left \{ t^{*}|l[t^{*}]=l[t], \frac{W}{2} \le t^{*}\le T-\frac{W}{2}\right \} $ and $\hat{\mathcal{P}}(t)=\left \{ t^{*}|l[t^{*}]\ne l[t], \frac{W}{2} \le t^{*}\le T-\frac{W}{2}\right \} 
$ denotes the indices of the frames with the same and different labels corresponding to the $t$-th frame, respectively.

Our NCA module aims at detecting the inter-neighbourhood consistency, where $N_{t}$ and $N_{t^{*}}$ tend to have similar feature distributions, whereas the distributions of $N_{t}$ and $N_{t^{'}}$ may be different. We define the problem of consistency detection as minimisation practice:   
\begin{equation}
\begin{split}
\begin{aligned}
\mathcal{L}_{nca}=&-\mathbb{E}_{N_{t} \sim \mathcal{X}(t)}\Big[\mathbb{E}_{N_{t^{*}} \sim \mathcal{N}(t)}\left[\log \mathcal{G}\left(N_{t},N_{t^{*}}\right)\right]
\\
&+\mathbb{E}_{N_{t^{'}} \sim \hat{\mathcal{N}}(t)}\left[ \log \left(1-\mathcal{G}\left(N_{t}, N_{t^{'}}\right)\right) \right]\Big] 
\end{aligned}
\label{equation:NCA_loss}
\end{split}
\end{equation} 
where $\mathbb{E}$ represents the expectation. $\mathcal{G}\left(N_{t},N_{t^{*}}\right)$ results in the probability of the neighbourhoods having similar feature distributions, formulated as follows:
\begin{equation}
\begin{split}
\begin{aligned}
\mathcal{G}(N_{t}, N_{t^{*}})= \textbf{G}([max(N_{t});max(N_{t^{*}})])
\end{aligned}
\label{equation:NCA_G}
\end{split}
\end{equation}
where $\textbf{G}$ is a MLP block (input dimension is $2D$, output dimension is 1) similar to $\textbf{S}$ in Eq. (\ref{equation:mlp}).
$[;]$ represents the concatenation operation. $max(\cdot)$ means max-pooling used to aggregating temporal features \cite{zhang2020semantics, sener2020temporal}. 
\begin{algorithm}[H]
\caption{The learning process of our SMC-NCA framework.}
\begin{algorithmic}[1]
\renewcommand{\algorithmicrequire}{\textbf{Input:}}
\renewcommand{\algorithmicensure}{\textbf{Output:}}
\REQUIRE Input features $\mathbf{V}$, temporal representation $\mathbf{X}$ and semantic representation $\mathbf{H}$. Line 2-Line 10: initial unsupervised learning. Line 12-Line 20: semi-supervised learning (stage 1). Line 21-Line 26: semi-supervised learning (stage 2). 
\STATE Generate sampled $\mathbf{V_{s}}$, $\mathbf{X_{s}}$ and $\mathbf{H_{s}}$
\FOR { epoch $e \leftarrow 1 $ to E1 }
\STATE  \footnotesize  $\mathcal{L}_{ap}^{P} \leftarrow -Sum \Big(\Big[\log\Big(\sigma(\mathbf{H}_{s}^{\top }\mathbf{X}_{s}/\xi)\Big)\Big]\odot \mathbf{I}\Big) / Sum( \mathbf{I})$

\STATE  \footnotesize $\textbf{M}  \leftarrow Cluster (\mathbf{V}_{s}, \mathbf{X}_{s}, \mathbf{H}_{s}) $ 

\STATE   \footnotesize  $\mathcal{L}_{ap}^{N} \leftarrow- Sum \Big(\Big[\log\Big(\sigma(-\mathbf{H}_{s}^{\top }\mathbf{X}_{s}/\xi   )\Big)\Big]\odot \mathbf{M}\Big) / Sum(\mathbf{M})$

\STATE   \footnotesize  $\mathcal{L}_{aa}^{N} \leftarrow- Sum\Big(\Big[\log\Big(\sigma(-\mathbf{H}_{s}^{\top }\mathbf{H}_{s}/\xi   )\Big)\Big]\odot \mathbf{M}\Big)  / Sum(\mathbf{M})$

\STATE  \footnotesize $\mathcal{L}_{pp}^{N} \leftarrow- Sum\Big(\Big[\log\Big(\sigma(-\mathbf{X}_{s}^{\top }\mathbf{X}_{s}/\xi   )\Big)\Big]\odot \mathbf{M}\Big)  / Sum(\mathbf{M})$

\STATE  \small $\mathcal{L}_{smc} \leftarrow \mathcal{L}_{ap}^{P}+\mathcal{L}_{ap}^{N}+\mathcal{L}_{aa}^{N}+\mathcal{L}_{pp}^{N}$ (on $\mathcal{D}$)
\STATE Minimize $\mathcal{L}_{smc}$  and update models $\mathbf{T}$ and $\mathbf{S}$
\ENDFOR
\STATE Obtain the best model evaluated by the linear classifier

\FOR { iter $i \leftarrow 1 $ to $ I $}
    \FOR { epoch $e \leftarrow 1 $ to E2 }
    \STATE  $\textbf{M}  \leftarrow$ GT
    \STATE \small $\mathcal{L}_{smc} \leftarrow\mathcal{L}_{ap}^{P}+\mathcal{L}_{ap}^{N}+\mathcal{L}_{aa}^{N}+\mathcal{L}_{pp}^{N}$ (on $\mathcal{D}_{L}$)
    \STATE Sample $ t \in \mathcal{Z}(t)=\left \{ t_{i}|1\le i \le K\right \}$
    \STATE $\mathcal{L}_{nca} \leftarrow NCA(t)$ 
    \STATE $\mathcal{L} \leftarrow\small \mathcal{L}_{smc}+\mathcal{L}_{nca} + \mathcal{L}_{ce} $ 
    \STATE Minimize $\mathcal{L}$ and update models $\mathbf{T}$,  $\mathbf{S}$, $\mathbf{G}$ and $\mathbf{C}$ 
    \ENDFOR
        \FOR { epoch $e \leftarrow $ 1  to E3}
        \STATE PL $\leftarrow \mathbf{C}(\mathcal{D}_{U})$
        \STATE  $\textbf{M}  \leftarrow$ PL $\cup$ GT
        \STATE $\mathcal{L} \leftarrow\small \mathcal{L}_{smc}+\mathcal{L}_{nca}$ (on $\mathcal{D}$) 
        \STATE Minimize $\mathcal{L}$ and update models $\mathbf{T}$,  $\mathbf{S}$ and $\mathbf{G} $ on $\mathcal{D}$
        \ENDFOR
\ENDFOR
\end{algorithmic}
\label{Alg:IAT}
\end{algorithm}

The loss function can be approximated as:
\begin{equation} 
\begin{split}
\begin{aligned}
\mathcal{L}_{nca}=&-\frac{1}{K\times M }\sum_{i=1}^{K}\sum_{j=1}^{M}\Big[\log \mathcal{G}(N_{t}^{i}, N_{t^{\ast }}^{j})+
\\
&\log(1- \mathcal{G}(N_{t}^{i},N_{t^{'}}^{j}))\Big]    
\end{aligned}
\label{equation:NCA_loss2}
\end{split}
\end{equation}

\begin{table*}[ht]
\centering
\caption{Component-wise analysis of the unsupervised representation learning framework with a linear classifier.  }
\label{tab:ablation_unsuper}
\begin{tabular}{m{4cm}<{\centering}|m{5.5cm}<{\centering}|m{5.5cm}<{\centering}}
\hline
\multirow{2}{*}{Method}                                                    &50Salads       &GTEA              \\\cline{2-3}
  & F1@\{10, 25, 50\} \qquad Edit \qquad Acc        & F1@\{10, 25, 50\} \qquad Edit \qquad Acc \\
 \hline
 \hline
$ \mathcal{L}_{ap}^{P}$($\textbf{M}_{in}$)+$ \mathcal{L}_{aa}^{N}$ & 36.8 \quad	31.1\quad	22.6\qquad	29.8\qquad57.1 &73.1 \quad	66.3\quad	48.5\qquad	65.5\qquad	70.2\\

$ \mathcal{L}_{ap}^{P}$($\textbf{I}$)+$ \mathcal{L}_{aa}^{N}$ &\textbf{51.9} \quad	\textbf{46.8}\quad	\textbf{38.2}\qquad	\textbf{39.7}\qquad	\textbf{72.4} & \textbf{75.8} \quad	\textbf{71.1}\quad	\textbf{53.0}\qquad	\textbf{68.0}\qquad	\textbf{73.8}\\
 \cline{2-3}
$ \mathcal{L}_{ap}^{P}$($\textbf{M}_{in}$)+$ \mathcal{L}_{ap}^{N}$ &39.1 \quad	33.8\quad	25.2\qquad	\textbf{31.8}\qquad	56.4 &69.8 \quad	63.7\quad	48.3\qquad	64.3\qquad	68.8\\

$ \mathcal{L}_{ap}^{P}$($\textbf{I}$)+$ \mathcal{L}_{ap}^{N}$ &\textbf{40.7} \quad	\textbf{35.6}\quad	\textbf{26.4}\qquad	30.9\qquad	\textbf{64.8} \quad &\textbf{73.0} \quad \textbf{67.6}\quad	\textbf{50.0}\qquad	\textbf{66.1}\qquad	\textbf{70.8}\\
\hline
\multicolumn{3}{c}{\textit{ Constructing positive pairs by $\textbf{M}_{in}$ (rely on the clustering results) or $I$ (without relying on the clustering) }}\\
\hline
$\mathcal{L}_{ap}^{P}$ + $ \mathcal{L}_{aa}^{N}$  & 51.9 \quad	46.8\quad	38.2\qquad	39.7\qquad	72.4 & 75.8 \quad	71.1\quad	53.0\qquad	68.0\qquad	73.8\\

$\mathcal{L}_{ap}^{P}$ + $ \mathcal{L}_{ap}^{N}$  & 40.7 \quad	35.6\quad	26.4\qquad	30.9\qquad	64.8 &73.0 \quad	67.6\quad	50.0\qquad	66.1\qquad	70.8\\

$\mathcal{L}_{ap}^{P}$ +  $ \mathcal{L}_{aa}^{N}$ + $ \mathcal{L}_{ap}^{N}$   & 55.2 \quad 50.1\quad	42.0\qquad	44.0\qquad	73.1 &76.2 \quad	\textbf{71.8}\quad	55.7\qquad	67.9\qquad	75.0\\

$\mathcal{L}_{ap}^{P}$ +  $ \mathcal{L}_{aa}^{N}$ + $ \mathcal{L}_{ap}^{N}$ + $ \mathcal{L}_{pp}^{N}$  & \textbf{56.5} \quad	\textbf{52.1}\quad	\textbf{42.8}\qquad	\textbf{45.4}\qquad	\textbf{74.3} &\textbf{77.4} \quad	71.2\quad	\textbf{56.1}\qquad	\textbf{69.0}\qquad	\textbf{75.7} \\
\hline
\multicolumn{3}{c}{\textit{Comparing different negative pairs }}\\
\hline
w/o dynamic clustering & 56.5 \quad	52.1\quad	42.8\qquad	45.4\qquad	74.3 &77.4 \quad	71.2\quad	56.1\qquad	69.0\qquad	75.7\\
w/ dynamic clustering & \textbf{58.1} \quad	\textbf{54.0}\quad	\textbf{43.5}\qquad	\textbf{46.3}\qquad	\textbf{75.1} &\textbf{78.9} \quad	\textbf{74.3}\quad	\textbf{59.2}\qquad	\textbf{73.0}\qquad	\textbf{76.2}\\
\hline
\multicolumn{3}{c}{\textit{Dynamic clustering facilitates contrastive learning  }}\\
\hline

\end{tabular}
\end{table*}

At this stage, our SMC is guided by ground-truth labels of labelled videos. Thus, the overall loss is defined as $\mathcal{L} = \mathcal{L}_{smc}+\mathcal{L}_{nca} + \mathcal{L}_{ce}$, where $\mathcal{L}_{ce}$ is the standard frame-level cross-entropy for action classification:
\begin{equation} 
\begin{split}
\begin{aligned}
\mathcal{L}_{ce}=-\frac{1}{T}\sum_{t=1}^{T}\sum_{a=1}^{A}y_{a} [t]\log(\hat{y}_{a}[t]) 
\end{aligned}
\label{equation:loss_ce}
\end{split}
\end{equation}
where $\hat{y}_{a}[t]$ represents the predicted probability of the $t$-th frame belonging to class a.
At the second stage, the ground-truth labels are combined with the pseudo labels (PL) generated by $\textbf{C}$ at the first stage to guide the learning of $\textbf{T}$, $\textbf{S}$ and $\textbf{G}$ for further updating frame-level representation. Our SMC-NCA framework is summarised in Algorithm \ref{Alg:IAT}.

\section{Experiments}
\subsection{Datasets and Evaluation}
\noindent  \textbf{Public Action Segmentation Datasets.} We evaluate the proposed method on three challenging datasets: \textbf{50Salads} \cite{stein2013combining} (50 videos, 19 actions), \textbf{GTEA}  \cite{fathi2011learning} (28 videos, 11 actions), and \textbf{Breakfast Actions}  \cite{kuehne2014language} (1712 videos, 10 complex activities, 48 actions). Following  \cite{singhania2022iterative}, we use the standard train-test splits for each dataset.

\noindent  \textbf{Mouse Social Behaviour Dataset.} Based on the dataset from our previous works \cite{zhou2021structured,jiang2021multi,9744557,zhou2022cross}, we introduce a new Parkinson’s Disease Mouse Behaviour (PDMB) dataset consisting of three groups of normal mice and three groups of mice with Parkinson’s Disease. It provides a valuable resource to study the behavioural patterns and characteristics of mice, particularly those with Parkinson’s Disease. More details can be found in Supplementary B.

\noindent \textbf{Evaluation.}  We follow the evaluation protocol used in ICC  \cite{singhania2022iterative} for both unsupervised and semi-supervised settings. In specific, for the public datasets, we use the same evaluation metrics taken for fully-supervised temporal action segmentation, including frame-wise accuracy (Acc), segmental Edit distance (Edit), and segmental F1 score at overlapping thresholds 10$\%$, 25$\%$, and 50$\%$ (denoted as F1@\{10, 25, 50\})  \cite{lea2017temporal}. The overlapping ratio is the intersection over union (IoU) ratio between the predicted and ground-truth action segments. We conduct cross-validation using the standard splits  \cite{singhania2022iterative}, and report the average. We evaluate the representation generated in unsupervised learning by training a linear classifier to classify frame-wise action labels. In the semi-supervised setting, we also report the average of 5 different selections to reduce the randomness brought by the training subset selection. For our PDMB dataset, we also use similar evaluation metrics and define 2 splits for cross-validation.

\subsection{Implementation Details.} 
All our experiments are performed on Nvidia Tesla P100 GPUs with 16GB memory. The parameters are optimised by the Adam algorithm. Learning Rate (LR), Weight Decay (WD), Epochs (Eps), and Batch Size (BS) used for our unsupervised and semi-supervised setups can be found in Tab. S2. Similar to \cite{singhania2022iterative}, the number of iterations is set to 4. Following the default of ICC \cite{singhania2022iterative}, we set the number of clusters in K-means to 40, 30 and 100 for 50Salads, GTEA and Breakfast datasets, respectively. At the beginning of unsupervised representation learning, $\xi$ is set to 1 for 50salads, GTEA and our PDMB dataset, and 0.1 for the large dataset Breakfast. In the NCA module, the length of a neighbourhood $W$, the number of the selected frames $K$, and the number of frames $M$ with the same label for each selected frame are set to 8, 1 and 10, respectively. The source code will be published after the paper has been accepted.

\subsection{Ablation Studies}

\subsubsection{Evaluation of Representation Learning}
\label{sec:section4.3.1}
\textbf{Effect of positive pairs construction.} As defined in Eq. (\ref{equation:P_ap}), we construct the positive pairs based on the inherent similarity between
frame-wise representations with temporal and semantic information rather than matrix, e.g., $\textbf{M}_{in}$ from the clustering outcomes. Tab. \ref{tab:ablation_unsuper} (Top) shows the impact of positive pair construction for the inter-information similarity learning on two public datasets (i.e., 50Salads and GTEA datasets). We replace $\textbf{I}$ in Eq. (\ref{equation:P_ap}) with $\textbf{M}_{in}$ to generate dense positive pairs, and the corresponding loss $\mathcal{L}_{ap}^{P}$($\textbf{M}_{in}$) is combined with $ \mathcal{L}_{aa}^{N}$ and $ \mathcal{L}_{ap}^{N}$ respectively for representation learning. As shown in Tab. \ref{tab:ablation_unsuper} (Top), we experience a significant decrease in performance using the contrastive learning guided by $\textbf{M}_{in}$, especially on the 50Salads dataset. This is likely caused by the fact that this design results in more clustering errors, thus affecting inter-information similarity learning.

\begin{table}[htb]
\setlength{\abovecaptionskip}{0cm} 
\setlength{\belowcaptionskip}{-0.0cm}
\centering
\caption{Ablation study on the Breakfast dataset.  }
\label{tab:ablation_unsuper_breakfast}
\begin{tabular}{m{3cm}<{\centering}|m{4.6cm}<{\centering}}
\hline
\multirow{2}{*}{Method}                                                    &Breakfast        \\\cline{2-2}
  & F1@\{10, 25, 50\} \qquad Edit \qquad Acc   \\
\hline
\hline
$\mathcal{L}_{ap}^{P}$ + $ \mathcal{L}_{aa}^{N}$  &57.4 \quad	52.6\quad	39.0\qquad	51.0\qquad	70.3 \\

$\mathcal{L}_{ap}^{P}$ + $ \mathcal{L}_{ap}^{N}$  &56.3 \quad	51.2\quad	38.4\qquad	50.3\qquad	69.1 \quad \\

$\mathcal{L}_{ap}^{P}$ +  $ \mathcal{L}_{aa}^{N}$ + $ \mathcal{L}_{ap}^{N}$   & 58.2 \quad 54.1\quad	41.0\qquad	52.1\qquad	71.7 \\

$\mathcal{L}_{ap}^{P}$ +  $ \mathcal{L}_{aa}^{N}$ + $ \mathcal{L}_{ap}^{N}$ + $ \mathcal{L}_{pp}^{N}$  & \textbf{59.0} \quad	\textbf{54.0}\quad	\textbf{41.9}\qquad	\textbf{52.2}\qquad	\textbf{71.9}  \\
\hline
w/o dynamic clustering & 59.0 \quad	54.0\quad	41.9\qquad	52.2\qquad	71.9 \\
w/ dynamic clustering & \textbf{59.7} \quad	\textbf{55.4}\quad	\textbf{42.8}\qquad	\textbf{52.7}\qquad	\textbf{72.1} \\
\hline
\end{tabular}
\end{table}

\begin{table}[ht]
\setlength{\abovecaptionskip}{0cm} 
\setlength{\belowcaptionskip}{-0.0cm}
\centering
\caption{ Comparing semantic feature extractors on the 50Salads dataset. }
\label{tab:comparison_mlp}
\begin{tabular}{m{2.5cm}<{\centering}|m{2.6cm}<{\centering}m{0.6cm}m{0.6cm}}
\hline
 Method & F1@\{10, 25, 50\} \quad &Edit \quad &Acc\\
\hline
\hline
MLP (1 layer)  & \textbf{58.1}\quad \textbf{54.0}\quad	\textbf{43.5}\qquad	&\textbf{46.3}\qquad	&\textbf{75.1}\\ 
MLP (3 layers) & 45.4\quad 41.0\quad	32.5\qquad	&37.3\qquad	&61.9\\
MLP (5 layers) & 39.4\quad 34.6\quad	25.1\qquad	&32.3\qquad	&57.6\\
Conv1d (1 layers) & 55.1\quad 50.0\quad	41.0\qquad	&44.1\qquad	&72.0\\
Conv1d (3 layers) & 42.7\quad 38.6\quad	28.1\qquad	&34.6\qquad	&59.3\\
 Conv1d (5 layers) & 37.9\quad 32.5\quad	22.9\qquad	&31.4\qquad	&54.1\\
\hline
\end{tabular}
\end{table}

\begin{figure*}[htb]
\begin{center}
\includegraphics[width=17cm]{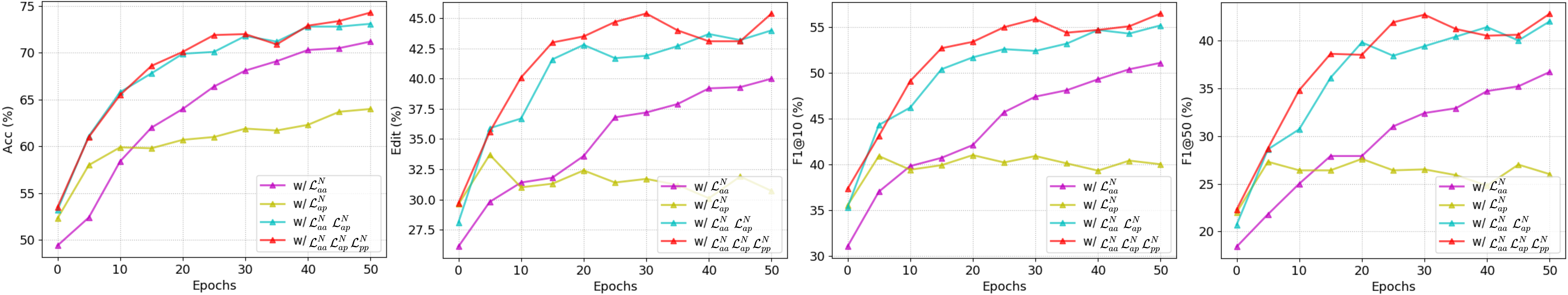}
\end{center}
\caption[Performance of unsupervised representation learning using different loss functions during training on 50Salads] {Performance of unsupervised representation learning using different loss functions during training on 50Salads. }
\label{fig:converge_curve}
\end{figure*}

\begin{figure*}[htb]
\begin{center}
\includegraphics[width=16cm]{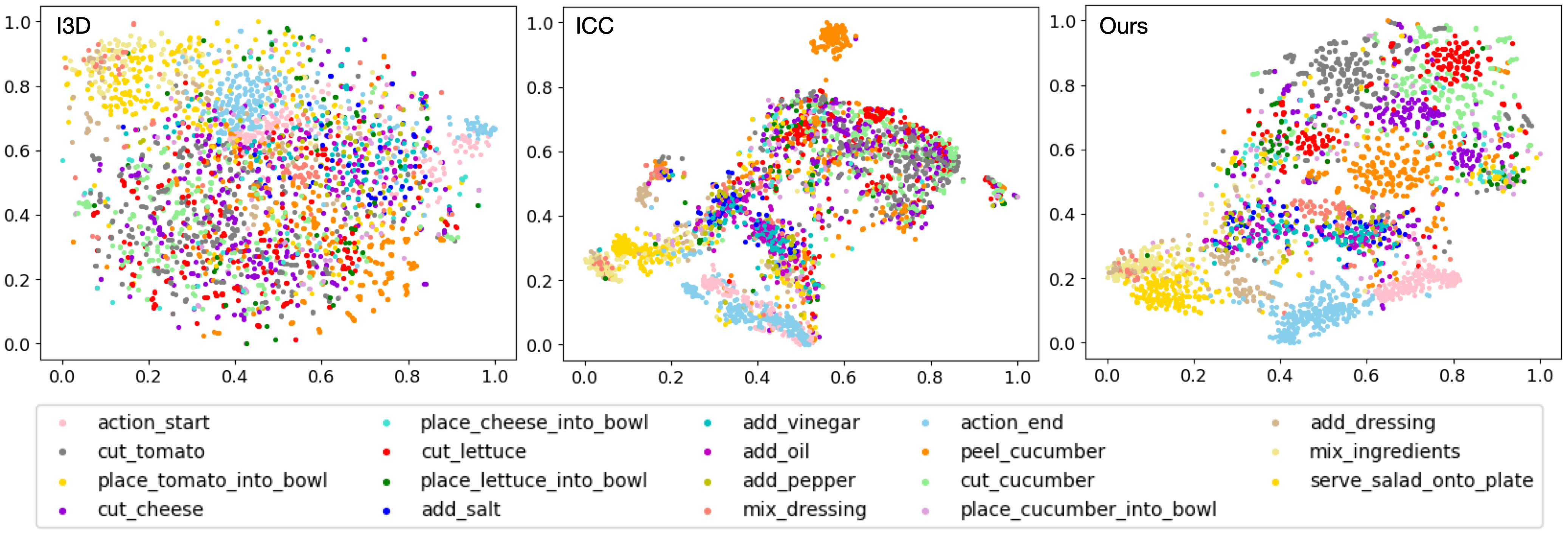}
\end{center}
\caption[t-SNE visualisation of the I3D feature and other features learned by ICC and our method on the 50Salads dataset] {t-SNE visualisation of the I3D feature and other features learned by ICC and our method. Each point represents an image frame. We show all action classes (19) of the 50Salads dataset in different colours.}
\label{fig:tsne_50salads}
\end{figure*}

\textbf{Effect of negative pairs construction.} Tab. \ref{tab:ablation_unsuper} (Middle) illustrates the results of the combinations of loss functions. In our SMC framework, we construct three types of dense negative pairs from the representations with temporal, semantic information and their combinations, respectively (the corresponding losses can be annotated as $ \mathcal{L}_{pp}^{N}$, $ \mathcal{L}_{aa}^{N}$ and $\mathcal{L}_{ap}^{N}$). As shown in Tab. \ref{tab:ablation_unsuper} (Middle), compared with $\mathcal{L}_{ap}^{N}$, $ \mathcal{L}_{aa}^{N}$ achieves better F1 and Edit scores and accuracy, with improvements of around 10$\%$ and 3$\%$ on the 50Salads and GTEA datasets. This demonstrates that contrasting learning using semantic representations enables effective dissimilarity learning between frames. In addition, using $ \mathcal{L}_{aa}^{N}$ with $\mathcal{L}_{ap}^{N}$ brings significant gains in all the metrics on the 50Salads and GTEA datasets, showing that these two losses are complementary to enhance contrastive learning as they impose intra- and inter-information dissimilarity learning, respectively. We further observe that the overall performance is slightly improved on both datasets by adding the contrastive loss with temporal representation, i.e., $ \mathcal{L}_{pp}^{N}$. 
On the Breakfast dataset, we can also achieve the best performance for all metrics by combining three types of negative pairs at the same time. The results are reported in Tab. \ref{tab:ablation_unsuper_breakfast}.
We also study the training process using different loss functions, as shown in Fig. \ref{fig:converge_curve}. As training progresses, the approach with three loss functions constantly outperforms the other approaches.

\textbf{Effect of dynamic clustering.} As described in Section \ref{sec:section3.3}, we exploit the dynamic matrix $\textbf{M}$ in Eq. (\ref{equation:kmeans}) to guide multi-level contrast. Tab. \ref{tab:ablation_unsuper} (Bottom) and  Tab. \ref{tab:ablation_unsuper_breakfast} (Bottom) show the effect of dynamic clustering. Note that the method without dynamic clustering refers to the approach with the fixed matrix  $\textbf{M}_{in}$. From  Tab. \ref{tab:ablation_unsuper} (Bottom) and Tab. \ref{tab:ablation_unsuper_breakfast} (Bottom), we discover that the segmentation performance can be further improved by applying the dynamic clustering to the selection of negative pairs. This is mainly because dynamic clustering allows us to further reduce the clustering errors by joint clustering on the temporal, semantic and pre-trained input features.

\begin{figure*}[htb]
\begin{center}
\includegraphics[width=17cm]{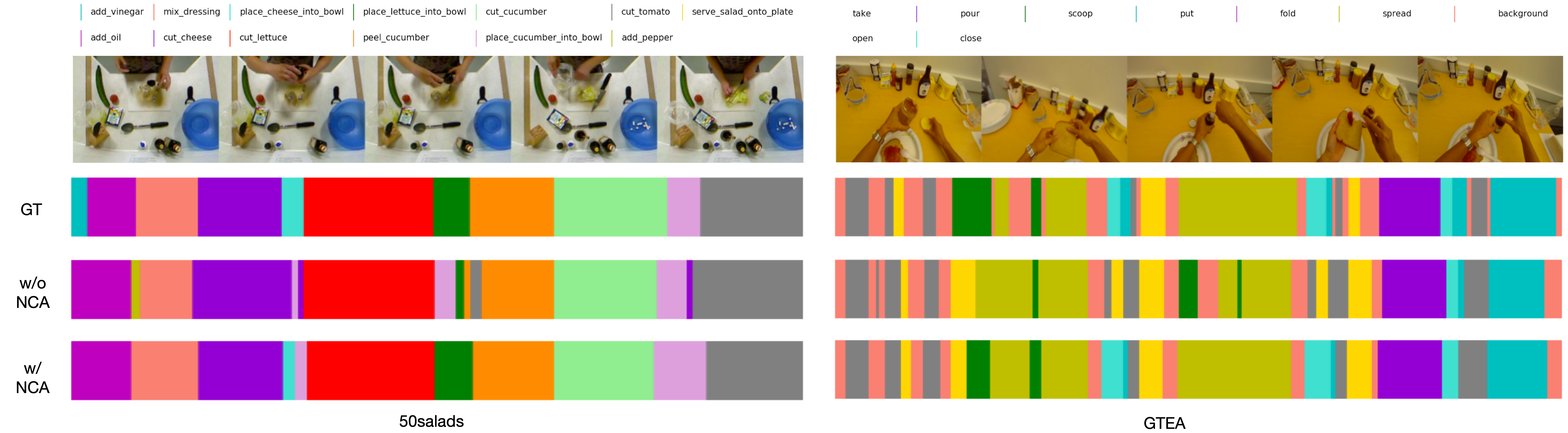}
\end{center}
\caption{Qualitative visualisation of segmentation results on the 50Salads and GTEA datasets with 5\% labelled data.}
\label{fig:NCA_all}
\end{figure*}

\begin{table*}[htb]
\centering
\caption{Comparing our proposed semi-supervised approach with the supervised learning approach using the same labelled data on the 3 benchmark datasets. }
\label{tab:ablation_semi}
\begin{tabular}{m{0.8cm}<{\centering}|m{1.8cm}<{\centering}|m{4.5cm}<{\centering}|m{4.5cm}<{\centering}|m{4.5cm}<{\centering}}
\hline
\multirow{2}{*}{$\%D_{L}$} & \multirow{2}{*}{Method} & 50salads & GTEA & Breakfast \\\cline{3-5}
& & F1@\{10, 25, 50\} \quad Edit \quad Acc & F1@\{10, 25, 50\} \quad Edit \quad Acc & F1@\{10, 25, 50\} \quad Edit \quad Acc\\
\hline
\hline
\multirow{3}{*}{5} &Supervised & 30.5\quad	25.4\quad  17.3\quad	26.3\quad	43.1  & 64.9\quad	57.5\quad 40.8\quad	59.2\quad	59.7  & 15.7\quad	11.8\quad	$\  $ 5.9\quad	19.8\quad	26.0 \\
                   &Semi-Super & 57.0\quad	53.1\quad	42.1\quad	48.9\quad	68.9      & 81.9\quad	79.3\quad 64.0\quad	77.9\quad	73.9     & 62.7\quad	57.0\quad 41.6\quad	60.2\quad 68.9\\
                   &Gain & \textbf{26.5\quad 27.7\quad 24.8\quad 22.6\quad 25.8}   & \textbf{17.0\quad 21.8\quad 23.2\quad 18.7\quad 14.2}   &  \textbf{47.0\quad  45.2\quad  35.7\quad  40.4\quad  42.9}\\ 
\hline
\multirow{3}{*}{10} &Supervised &45.1\quad 38.3\quad 26.4\quad 38.2\quad 54.8   &66.2\quad 61.7\quad 45.2\quad 62.5\quad 60.6    & 35.1\quad 30.6\quad 19.5\quad 36.3\quad 40.3\\
                   &Semi-Super & 70.3\quad	66.3\quad	54.7\quad	61.3\quad	73.6   &87.7\quad	84.2\quad	71.7\quad	83.3\quad	77.5      &64.8\quad	 60.2\quad	45.0\quad	63.4\quad	69.7  \\
                   &Gain & \textbf{25.2\quad 28.0\quad 28.3\quad 23.1\quad 18.8} & \textbf{21.5\quad 22.5\quad 26.5\quad 20.8\quad 16.9}    &\textbf{29.7\quad 29.6\quad 25.5\quad 27.1\quad 29.4} \\
\hline          
\multirow{3}{*}{100} &Supervised & 75.8\quad 73.1\quad 62.3\quad 68.8\quad 79.4    &90.1\quad 87.8\quad 74.9\quad 86.7\quad 79.5     & 69.4\quad 65.9\quad 55.1\quad 66.5\quad 73.4\\
                   &Semi-Super &86.9\quad	84.8\quad	78.9\quad	80.7\quad	87.0      & 92.7\quad	91.0\quad	81.5\quad	88.3\quad	82.6      & 73.8\quad	69.7\quad	56.8\quad	70.9\quad	76.4  \\
                   &Gain & \textbf{11.1\quad 11.7\quad 16.6\quad 11.9\quad $\ $7.6}    
                   & \textbf{$\ $2.6\quad  $\ $3.2\quad $\ $6.6\quad $\ $1.6\quad $\ $3.1}   
                   &$\ $\textbf{4.4\quad $\  $3.8\quad $\ $1.7\quad $\ $4.4\quad $\ $3.0} \\
\hline

\end{tabular}
\end{table*}

\textbf{Effect of different semantic feature extractors.} Tab. \ref{tab:comparison_mlp} shows the results of using different semantic feature extractors for constructing multi-level contrastive learning. 
Semantic features provide context about the scene or objects present in the video frames, which highlights action-specific attributes. First, they assist in creating inter-information positive pairs to reinforce intra-class similarity. For instance, frames at corresponding positions in the sequence, both in terms of semantic and temporal features, are expected to belong to the same action category. Second, they contribute to enlarging inter-class dissimilarity by constructing inter-information negative pairs and intra-information negative pairs, respectively. 'MLP (1 layer)' means a single-layer MLP, which has one hidden layer between the input and output layers. 'Conv1d' refers to a one-dimensional convolutional layer. We can observe that greater complexity of semantic feature extractors may lead to worse performance. Therefore, we choose MLP (1 layer) as the semantic feature extractor.

\begin{table}
\centering
\caption{Performance of the NCA module on the 50Salads, GTEA and Breakfast datasets (5$\%$).}
\label{tab:ablatio_NCA}
\begin{tabular}{m{1.2cm}<{\centering}|m{1.5cm}<{\centering}|m{2.4cm}<{\centering}m{0.6cm}m{0.6cm}}      \hline
Dataset & Method & F1@\{10, 25, 50\} \quad &Edit \quad &Acc\\
\hline
\hline
\multirow{3}{*}{50Salads} & w/o $\mathcal{L}_{nca}$ &52.1\quad 47.0\quad  35.8 &43.9  &67.5\\
& w/ $\mathcal{L}_{nca}$ & 57.0\quad 53.1\quad	42.1\qquad	&48.9\qquad	&68.9\\
& Gain &\textbf{ 4.9}\quad \textbf{ 6.1}\quad \textbf{ 6.3} \qquad  &\textbf{ 5.0} &\textbf{ 1.4}\\ 
\hline
\multirow{3}{*}{GTEA} & w/o $\mathcal{L}_{nca}$ &77.5\quad	74.2\quad	57.7	&71.8	&71.4 \\
& w/ $\mathcal{L}_{nca}$ & 81.9\quad	79.3\quad64.0	&77.9	&73.9 \\
& Gain  &\textbf{ 4.4}\quad	\textbf{$\ $ 5.1}\quad	\textbf{ 6.3}	\qquad &\textbf{ 6.1}	&\textbf{ 2.5}\\
\hline
\multirow{4}{*}{\makecell[c]{Breakfast}} & w/o $\mathcal{L}_{nca}$ & 60.8\quad	54.8\quad	39.5	&58.1 &68.1 \\
& w/ $\mathcal{L}_{nca}$& 62.7\quad	57.0\quad	41.6 	&60.2 &68.9 \\
& Gain  &\textbf{ 1.9} \quad	\textbf{ 2.2} \quad \textbf{1.1}	\qquad &\textbf{ 2.1}	&\textbf{ 0.8}\\ 
\hline
\end{tabular}
\end{table}

\textbf{Qualitative Results.} In this section, we employ t-Distributed Stochastic Neighbor Embedding (t-SNE) to visualise the I3D feature alongside other features learned by ICC and our proposed method on the 50Salads dataset, as shown in Fig. \ref{fig:tsne_50salads}. Each colour in the figure corresponds to a distinct action. The features selected for visualisation are randomly sampled from the combined multi-resolution representation extracted by the C2F-TCN model \cite{singhania2022iterative}. Notably, our method exhibits superior class separation, indicating the strong representation learning capability facilitated by our Semantic-guided Multi-level Contrast scheme. This enhanced separation shows the effectiveness of our approach in capturing more discriminative representation for action segmentation.

\begin{table}
\centering
\caption{ Comparing our proposed unsupervised representation learning approach with other existing methods on the 3 benchmarks. }
\label{tab:comparison_unsuper}
\begin{tabular}{m{1.2cm}<{\centering}|m{1.5cm}<{\centering}|m{2.8cm}<{\centering}m{0.6cm}m{0.6cm}}
\hline
Dataset & Method & F1@\{10, 25, 50\} \quad &Edit \quad &Acc\\
\hline
\hline
\multirow{4}{*}{50Salads} & I3D &12.2\quad $\ $7.9\quad $\ $4.0 &$\ $8.4 &55.0\\
& ICC\cite{singhania2022iterative} & 40.8\quad 36.2\quad 28.1\qquad &32.4\qquad &62.5 \\
& SMC & 58.1\quad 54.0\quad	43.5\qquad	&46.3\qquad	&75.1\\& Gain &\textbf{17.3}\quad \textbf{17.8}\quad \textbf{15.4}\quad & \textbf{13.9} &\textbf{12.6}\\ 
\hline
\multirow{4}{*}{GTEA} & I3D &48.5\quad 42.2\quad 26.4\quad &40.2 & 61.9 \\
& ICC \cite{singhania2022iterative} &70.8\quad 65.0\quad 48.0 & 65.7& 69.1 \\
& SMC & 78.9\quad	74.3\quad	59.2	&73.0	&76.2 \\
& Gain  &\textbf{$\ $8.1}\quad 	\textbf{$\ $9.3}\quad	\textbf{11.2}	&\textbf{$\ $7.3}	&\textbf{$\ $7.1}\\
\hline
\multirow{4}{*}{Breakfast} & I3D  & $\ $4.9\quad $\ $2.5\quad  $\ $0.9 &$\ $5.3 &30.2 \\
& ICC\cite{singhania2022iterative} & 57.0\quad 51.7\quad 39.1 &51.3 &70.5 \\
& SMC & 59.7\quad 55.4\quad 42.8 & 52.7 & 72.1 \\
& Gain &\textbf{$\ $2.7}\quad \textbf{$\ $3.7}\quad \textbf{$\ $3.7}	&\textbf{$\ $1.4}	&\textbf{$\ $1.6}\\
\hline
\end{tabular}
\end{table}

\subsubsection{Evaluation of Semi-supervised Learning}
\label{sec:section4.3.2}
\noindent\textbf{Effect of the NCA unit.} In Section \ref{sec:section3.4}, we propose the NCA module to alleviate over-segmentation issues when only a fraction of videos are used for semi-supervised learning. Tab. \ref{tab:ablatio_NCA} shows the results where only 5$\%$ of labelled videos are used for semi-supervised learning on the 50Salads, GTEA and Breakfast datasets. While the baseline without $\mathcal{L}_{nca}$ achieves the expected frame-wise accuracy, it suffers from severe over-segmentation as indicated by the low F1 and Edit scores. By incorporating $\mathcal{L}_{nca}$ within the SMC module, we observe a significant performance boost, up to 6.1$\%$ improvement for the Edit score, and up to 6.3$\%$ improvement for the F1 score on the 50 Salads and GTEA. 
These results show that our proposed NCA module is capable of reducing over-segmentation errors by identifying the spatial consistency between the action segments.

\begin{figure*}
\begin{center}
\includegraphics[width=16cm]{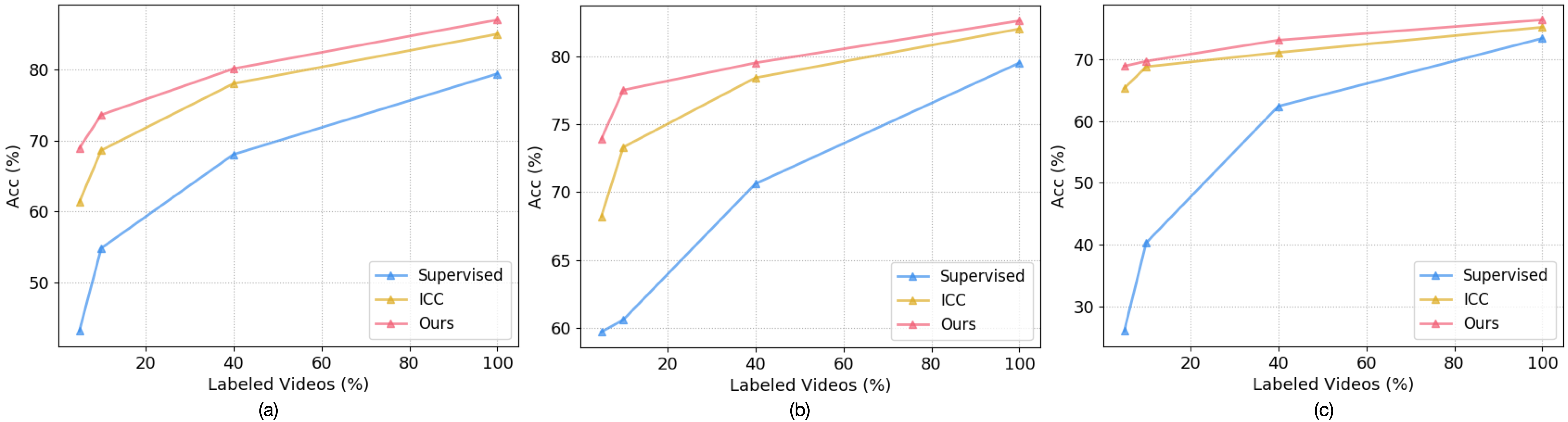}
\end{center}
\caption{Segmentation accuracy of different methods on (a) 50Salads, (b) GTEA and (c) Breakfast datasets. Our semi-supervised approach continuously achieves satisfactory performance over semi-supervised (5$\%$, 10$\%$ and 40$\%$) and fully-supervised (100$\%$) setups.}
\label{fig:comp_semi_all}
\end{figure*}

\begin{table*}
\centering
\caption{Comparing our proposed semi-supervised approach with ICC on the 3 benchmarks.  }
\label{tab:comparison_allmetrics}
\begin{tabular}{m{0.8cm}<{\centering}|m{2.2cm}<{\centering}|m{4.4cm}<{\centering}|m{4.4cm}<{\centering}|m{4.4cm}<{\centering}}
\hline
\multirow{2}{*}{$\%D_{L}$} & \multirow{2}{*}{Method} & 50salads & GTEA & Breakfast \\\cline{3-5}
& & F1@\{10, 25, 50\} \quad Edit \quad Acc & F1@\{10, 25, 50\} \quad Edit \quad Acc & F1@\{10, 25, 50\} \quad Edit \quad Acc\\
\hline
\hline
\multirow{3}{*}{5} &ICC \cite{singhania2022iterative} & 52.9\quad 49.0\quad 36.6\quad 45.6\quad 61.3  & 77.9\quad	71.6\quad 54.6\quad	71.4\quad	68.2  & 60.2\quad	53.5\quad	35.6\quad	56.6\quad	65.3 \\
                   &Ours (SMC-NCA) & 57.0\quad	53.1\quad	42.1\quad	48.9\quad	68.9      & 81.9\quad	79.3\quad 64.0\quad	77.9\quad	73.9     & 62.7\quad	57.0\quad 41.6\quad	60.2\quad 68.9\\
                   &Gain &  \textbf{$\ $4.1\quad	 $\ $4.1\quad	 $\ $5.5\quad	 $\ $3.3\quad	 $\ $7.6} &  \textbf{$\ $4.0\quad	 $\ $7.7\quad	 $\ $9.4\quad	 $\ $6.5\quad	 $\ $5.7} &  \textbf{$\ $2.5\quad	 $\ $3.4\quad	 $\ $6.0\quad	 $\ $3.6\quad	 $\ $3.6}\\ 
\hline
\multirow{3}{*}{10} &ICC \cite{singhania2022iterative} &67.3\quad 64.9\quad 49.2\quad 56.9\quad 68.6  &83.7\quad  81.9\quad  66.6\quad 76.4\quad 73.3    & 64.6\quad 59.0\quad 42.2\quad 61.9\quad 68.8\\
                   &Ours (SMC-NCA) & 70.3\quad	66.3\quad	54.7\quad	61.3\quad	73.6   &87.7\quad	84.2\quad	71.7\quad	83.3\quad	77.5      &64.8\quad	 60.2\quad	45.0\quad	63.4\quad	69.7  \\
                   &Gain &  \textbf{$\ $3.0\quad	 $\ $1.4\quad	 $\ $5.5\quad	 $\ $4.4\quad	 $\ $5.0} &  \textbf{$\ $4.0\quad	 $\ $2.3\quad	 $\ $5.1\quad	 $\ $6.9\quad	 $\ $4.2} &  \textbf{$\ $0.2\quad	 $\ $1.2\quad	 $\ $2.8\quad	 $\ $1.5\quad	 $\ $0.9} \\
\hline  

\multirow{3}{*}{40} &ICC \cite{singhania2022iterative}& \quad $\backslash$\qquad   $\backslash$\qquad   $\backslash$\qquad   $\backslash$\qquad  78.0  & \quad $\backslash$\qquad   $\backslash$\qquad   $\backslash$\qquad   $\backslash$\qquad 78.4    & \quad $\backslash$\qquad   $\backslash$\qquad   $\backslash$\qquad   $\backslash$\qquad 71.1\\
                   &Ours (SMC-NCA) & 76.5\quad	73.6\quad	65.6\quad	67.9\quad	80.1  &85.8\quad	83.6\quad	72.8\quad	80.2\quad	79.5     &69.0\quad	64.2\quad	49.5\quad	67.1\quad	73.1  \\
                   &Gain & \quad $\backslash$\qquad   $\backslash$\qquad  $\backslash$\qquad   $\backslash$\qquad $\ $\textbf{2.1} &\quad $\backslash$\qquad   $\backslash$\qquad   $\backslash$\qquad   $\backslash$\qquad  $\ $\textbf{1.1} &\quad $\backslash$\qquad   $\backslash$\qquad   $\backslash$\qquad   $\backslash$\qquad $\ $\textbf{2.0}\\
\hline    

\multirow{3}{*}{100} &ICC \cite{singhania2022iterative} & 83.8\quad 82.0\quad 74.3\quad 76.1\quad 85.0    &91.4\quad 89.1\quad 80.5\quad 87.8\quad 82.0     & 72.4\quad 68.5\quad 55.9\quad 68.6\quad 75.2\\
                   &Ours &86.9\quad	84.8\quad	78.9\quad	80.7\quad	87.0      & 92.7\quad	91.0\quad	81.5\quad	88.3\quad	82.6      & 73.8\quad	69.7\quad	56.8\quad	70.9\quad	76.4  \\
                   &Gain & \textbf{$\ $3.1\quad	 $\ $2.8\quad	 $\ $4.6\quad	 $\ $4.6\quad	 $\ $2.0}
                   &  \textbf{$\ $1.3\quad	 $\ $1.9\quad	 $\ $1.0\quad	 $\ $0.5\quad	 $\ $0.6 }
                   &  \textbf{$\ $1.4\quad	 $\ $1.2\quad	 $\ $0.9\quad	 $\ $2.3\quad	 $\ $1.2}\\
\hline

\end{tabular}
\end{table*}

\textbf{Semi-supervised \textit{Vs} Supervised.} We also compare our proposed semi-supervised approach with the supervised counterpart using the same labelled data on the 3 benchmarks, and the results are reported in Tab. \ref{tab:ablation_semi}. Performance improvement is noticeable over all the evaluation metrics in both semi-supervised (5$\%$ and 10$\%$) and fully-supervised (100$\%$) setups, confirming the effectiveness of our semi-supervised learning framework.

\textbf{Qualitative Results.}  Qualitative results of the NCA module are presented in Fig. \ref{fig:NCA_all}. When observing prediction results without the NCA module, we notice certain over-segmentation errors characterised by numerous small segments. However, our NCA effectively alleviates this issue by identifying spatial consistency between neighbourhoods centered at different frames. This demonstrates that our proposed NCA ensures smoother and more coherent segmentation results.

\subsection{Comparison against other State-of-the-Art Approaches}
In this section, we compare our proposed unsupervised representation learning and semi-supervised learning framework against the other existing methods on three public datasets, i.e., 50 Salads, GTEA and Breakfast datasets and our PDMB dataset. Tabs. \ref{tab:comparison_unsuper} and \ref{tab:comparison_sota} report the results of the unsupervised representation learning and supervised learning with different levels (i.e., fully-, weakly- and semi-supervised) on the three human action datasets. Tabs. \ref{tab:comparison_unsuper_pdmb} and \ref{tab:comparison_sota_pdmb} show the results on our mouse social behaviour dataset.

\subsubsection{Comparison on human action datasets}
As shown in Tab. \ref{tab:comparison_unsuper}, our SMC is superior to the standard ICC in all the metrics, up to 17.8$\%$ gain for F1 score, 13.9$\%$ gain for Edit score and 12.6$\%$ gain for the frame-wise accuracy on the 50 Salads dataset. We observe a similar tendency on the GTEA dataset, with improvements of 11.2$\%$, 7.3$\%$ and 7.1$\%$ for the F1 score, Edit score and accuracy, respectively. The performance of our SMC is primarily attributed to its ability to capture and exploit intra- and inter-information variations. Regarding the Breakfast dataset with complex activities, we keep the video-level contrastive loss in ICC \cite{singhania2022iterative} and combine it with our SMC to ensure a fair comparison. The results in Tab. \ref{tab:comparison_unsuper} show that our approach is marginally better than ICC for all the evaluation metrics.

\begin{table}
\centering
\caption{Segmentation accuracy comparison against the state-of-the-art methods under various supervisions on 3 benchmark datasets. }
\label{tab:comparison_sota}
\begin{tabular}{m{1cm}<{\centering}|m{3.2cm}<{\centering}|m{0.8cm}<{\centering}m{0.8cm}<{\centering}m{0.8cm}<{\centering}}
\hline
  & Method & 50Salads  &GTEA &Breakfast\\
\hline
\hline
\multirow{7}{*}{Fully} & MSTCN\cite{farha2019ms} &83.7  &78.9 &67.6\\
                          & SSTDA \cite{chen2020action} &83.2 &79.8 & 70.2\\
                          & C2F-TCN \cite{singhania2021coarse} &79.4  &79.5 & 73.4\\
                          & ASFormer \cite{yi2021asformer} &85.6  &79.7 & 73.5\\
                          & CETNet \cite{wang2022cross} &86.9  &80.3 & 74.9\\
                           & DiffAct \cite{liu2023diffusion} & \textbf{88.9} & 82.2 & 76.4\\
                          & ICC \cite{singhania2022iterative} (100$\%$) &85.0  &82.0 & 75.2\\
                          & Ours (SMC-NCA)(100$\%$) &87.0 &\textbf{82.6} & \textbf{76.4} \\
\hline                         
\multirow{2}{*}{Weakly} & SSTDA \cite{chen2020action} (65$\%$) &80.7  &75.7 &65.8\\
                          & Timestamp \cite{li2021temporal} &75.6 &66.4 & 64.1\\
\hline                                        
\multirow{6}{*}{Semi}     & ICC \cite{singhania2022iterative} (40$\%$) &78.0  &78.4 &71.1\\
                          & Ours (SMC-NCA)(40$\%$) &\textbf{80.1}  &\textbf{79.5} &\textbf{73.1}\\ \cline{2-5}
                          & ICC \cite{singhania2022iterative} (10$\%$) &68.6 &73.3 & 68.8\\
                          & Ours (SMC-NCA)(10$\%$) &\textbf{73.6} &\textbf{77.5} & \textbf{69.7}\\ \cline{2-5}
                          & ICC \cite{singhania2022iterative} (5$\%$) &61.3 &68.2 & 65.3\\
                          & Ours (SMC-NCA)(5$\%$) &\textbf{68.9} &\textbf{73.9} & \textbf{68.9}\\
\hline                           
\end{tabular}
\end{table}

\begin{table}[htb]
\setlength{\abovecaptionskip}{0cm} 
\setlength{\belowcaptionskip}{-0.0cm}
\centering
\caption{ Comparisons against other semi-supervised learning methods on 5$\%$ labelled data. }
\label{tab:comparison_ficmatch}
\begin{tabular}{m{1cm}<{\centering}|m{2.2cm}<{\centering}|m{2.6cm}<{\centering}m{0.5cm}m{0.5cm}}
\hline
Dataset & Method & F1@\{10, 25, 50\} \quad &Edit \quad &Acc\\
\hline
\hline
\multirow{3}{*}{50Salads}  & FixMatch \cite{sohn2020fixmatch} &47.0\quad 41.3\quad 28.3\quad &38.3\quad &53.3\\
& UniMatch \cite{yang2023revisiting}  & 50.3\quad	45.9\quad	30.4	&39.7	&54.4 \\ 
& Ours (SMC-NCA)  & \textbf{57.0}\quad	\textbf{53.1}\quad	\textbf{42.1}	&\textbf{48.9}	&\textbf{68.9} \\ 
\hline
\multirow{3}{*}{GTEA} & FixMatch \cite{sohn2020fixmatch} & 69.1\quad  63.9\quad  44.1\quad & 59.5 & 62.0\\
& UniMatch \cite{yang2023revisiting}  & 65.4\quad 60.0\quad	40.8\qquad	&57.4\qquad	&58.7\\ 
& Ours (SMC-NCA)  & \textbf{81.9}\quad \textbf{79.3}\quad	\textbf{64.0}\qquad	&\textbf{77.9}\qquad	&\textbf{73.9}\\ 
\hline
\end{tabular}
\end{table}

In Tab. \ref{tab:comparison_sota}, we show the implementation results of our semi-supervised method using 100$\%$ labelled data, the best frame-wise accuracy among all the fully-supervised approaches. For the semi-supervised setting, we literally perform unsupervised representation learning and semi-supervised classification for 5$\%$, 10$\%$ and 40$\%$ of the training data, and the results show that our framework consistently outperforms ICC by a significant margin. It is noticed that the performance gap (i.e., 7.6$\%$, 5.7$\%$ and 3.6$\%$ improvements in accuracy on the three datasets) between the two methods using only 5$\%$ labelled videos is larger than that using more labelled data. This confirms that our approach is capable of dealing with semi-supervised learning problems in the presence of a small amount of labelled data. As shown in Tab. \ref{tab:comparison_allmetrics} and Fig. \ref{fig:comp_semi_all}, we evaluate our method on three benchmarks and achieve continuously better results for all the performance measures.

We have compared our method with ICC (the most relevant to our work) in Tab. \ref{tab:comparison_sota}. Regarding another study \cite{ding2022leveraging} on semi-supervised temporal action segmentation, since it does not provide publicly available code and implementation details, we are unable to directly compare it with our method. As mentioned in \cite{singhania2022iterative}, ICC is the first work for semi-supervised action segmentation and is not directly comparable to other works. Most existing image-based semi-supervised learning techniques are not suitable for action segmentation. The reason is that data augmentations (e.g., rotation and transformation) should be applied to input images, but it is non-trivial for action segmentation as the inputs are pre-computed feature vectors \cite{ding2022leveraging}. However, we managed to compare our method against two popular semi-supervised learning methods, i.e., FixMatch \cite{sohn2020fixmatch} and UniMatch \cite{yang2023revisiting}. In order to reproduce these methods, we designed different data augmentation strategies (e.g., noise) to generate input features with different levels of augmentations. Specifically, for FixMatch, we apply scale transformation (0.2 to 2) for weak augmentation and add Gaussian noise (mean 0, standard deviation 1) \cite{tarvainen2017mean} for strong augmentation. Similarly, for UniMatch, we use the same approach for weak augmentation and adopt channel dropout with a probability of 50\% as our feature perturbation \cite{yang2023revisiting}. Additionally, we add different Gaussian noises (standard deviation 1 and 1.5) for strong augmentations. The results reported in Tab.  \ref{tab:comparison_ficmatch} further verify the superiority of our method.

\subsubsection{Computational complexity} To analyse the computational complexity of our proposed SMC-NCA framework,  we report the FLOPs and parameter numbers on the 50Salads dataset using an input with the shape $1\times 2048\times960$ (1, 2048 and 960 denote bath size, feature dimension and temporal length), as shown in  Tab. \ref{tab:comparison_flops}. Specifically, we investigate the computational complexity of three distinct base TCN models, namely ED-TCN \cite{lea2017temporal} (an encoder-decoder architecture) MS-TCN \cite{farha2019ms} (we only use the single-stage configuration as MS-TCN is not designed for representation learning \cite{singhania2022iterative}) and C2F-TCN \cite{singhania2023c2f,singhania2022iterative}. To perform semi-supervised learning, we apply the learning strategies proposed in \cite{singhania2022iterative} to these models and the results are reported in the first three rows. We can observe that MS-TCN has the fewest parameters and FLOPs, but its segmentation performance is the poorest. One of the reasons can be that MS-TCN does not have multiple temporal resolution representations like encoder-decoder architecture which is important for unsupervised representation learning. Notably, ICC leveraging C2F-TCN emerges as the top performer among the three methods, demonstrating superior segmentation performance at a computational expense less than half of that of ED-TCN.

As mentioned in Sec. \ref{sec:section3}, we construct a novel SMC scheme to boost unsupervised representation learning by introducing a simple semantic feature extractor (single-layer MLP in Tab . \ref{tab:comparison_mlp}). By applying our SMC to the C2F-TCN backbone of ICC, our proposed approach demonstrates higher computational demands compared to ICC. Specifically, our method with SMC exhibits a notable increase in both FLOPs (from 2.4G to 5.4G) and parameter numbers (from 4.3M to 7.4M).  This can be attributed to the addition of the MLP that maps input \textbf{V} of feature dimension E (i.e., 2048) to output \textbf{H} of feature dimension D (i.e., 1536). Despite the increased computational cost, our approach achieves superior performance in both unsupervised and semi-supervised settings. Moreover, incorporating the NCA module slightly increases the parameter count by 0.4M while maintaining the same FLOPs (5.4G). 
However, the overall semi-supervised performance is significantly improved compared to ICC. It is noticeable that our proposed SMC and NCA are exclusively utilised during model training to enhance the semi-supervised segmentation performance of the backbone. Consequently, they do not contribute to any increase in inference time per video.

\begin{table}
\centering
\caption{Analysis of computational cost on the 50Salads datasets in terms of FLOPs and parameter number (Params). The semi-supervised results of Edit and Acc are based on 5\% labelled data.}
\label{tab:comparison_flops}
\begin{tabular}{c|c|c|cc|cc}
\hline
 \multirow{2}{*}{Method} & \multirow{2}{*}{FLOPs} & \multirow{2}{*}{Params} & \multicolumn{2}{c|}{Unsupervised} &  \multicolumn{2}{c}{Semi} 
\\ \cline{4-7} 
&  &  & Edit & Acc & Edit & Acc\\ 
\hline
\hline
ED-TCN \cite{lea2017temporal} & 7.3G & 8.2M &  $\backslash$ &  $\backslash$ & 32.7 & 46.4 \\
MS-TCN \cite{farha2019ms} & \textbf{0.3G} & \textbf{0.3M }&  $\backslash$ &  $\backslash$ & 23.4 & 38.3 \\
ICC \cite{singhania2022iterative}  & 2.4G &4.3M &32.4 &62.5
& 45.6 & 61.3\\
Ours (SMC) & 5.4G &7.4M &\textbf{46.3}\qquad	&\textbf{75.1} & 43.9 &67.5 \\
Ours (SMC-NCA) & 5.4G &7.8M &  $\backslash$ &  $\backslash$ & \textbf{48.9} & \textbf{68.9}\\
\hline
\end{tabular}
\end{table}

\begin{table}
\centering
\caption{ Comparing our proposed unsupervised representation learning approach with ICC on the PDMB dataset. }
\label{tab:comparison_unsuper_pdmb}
\begin{tabular}{m{1.2cm}<{\centering}|m{1.5cm}<{\centering}|m{2.8cm}<{\centering}m{0.6cm}m{0.6cm}}
\hline
Dataset & Method & F1@\{10, 25, 50\} \quad &Edit \quad &Acc\\
\hline
\hline
\multirow{3}{*}{PDMB}
& ICC \cite{singhania2022iterative} & 37.5\quad 32.5\quad 20.6\qquad &35.2\qquad &40.1 \\
& SMC & 61.1\quad 59.6\quad	47.2\qquad	&55.3\qquad	&62.5\\& Gain &\textbf{23.6}\quad \textbf{27.1}\quad \textbf{26.6}\quad & \textbf{20.3} &\textbf{22.4}\\ 
\hline
\end{tabular}
\end{table}

\begin{table}
\centering
\caption{Comparing our proposed framework with ICC on the PDMB dataset. }
\label{tab:comparison_sota_pdmb}
\begin{tabular}{m{0.8cm}<{\centering}|m{2.2cm}<{\centering}|m{4.2cm}<{\centering}}
\hline
\multirow{2}{*}{$\%D_{L}$} & \multirow{2}{*}{Method} & PDMB \\\cline{3-3}
& & F1@\{10, 25, 50\} \quad Edit \quad Acc\\
\hline
\hline
\multirow{3}{*}{10} &ICC\cite{singhania2022iterative} & 50.0\quad	42.5\quad  25.5\quad	49.7\quad	44.5 \\
                   &Ours (SMC-NCA) & \textbf{62.0}\quad	\textbf{57.1}\quad	\textbf{40.7}\quad	\textbf{52.7}\quad	\textbf{55.1}    \\
\hline
\multirow{3}{*}{50} &ICC\cite{singhania2022iterative} &63.5\quad 59.5\quad 40.5\quad 54.4\quad 58.6  \\
                   &Ours (SMC-NCA) & \textbf{68.0}\quad	\textbf{63.3}\quad	\textbf{46.6}\quad	\textbf{59.0}\quad	\textbf{66.8}   \\
\hline          
\multirow{3}{*}{100} &ICC\cite{singhania2022iterative} & 68.1\quad 63.1\quad 46.1\quad 59.8\quad 69.4  \\
                   &Ours (SMC-NCA) &\textbf{71.7}\quad	\textbf{67.5}\quad	\textbf{51.7}\quad	\textbf{63.0}\quad	\textbf{73.6}    \\
\hline 
\end{tabular}
\end{table}

\subsubsection{Comparison on mouse social behaviour dataset}
Temporal dependency modelling is crucial for the detection and segmentation of human actions in long videos  \cite{farha2019ms,yi2021asformer}. For animal behaviours such as mouse social behaviour recognition in long videos, behavioural correlations are also of importance to infer the social behaviour label of each frame  \cite{jiang2021multi}. Therefore, to demonstrate the generalisation and effectiveness of our proposed framework, we conducted experiments on our PDMB dataset. Tab. \ref{tab:comparison_unsuper_pdmb} depicts that our proposed SMC outperforms ICC by a large margin, offering an improvement of more than 20$\%$ for all metrics on the PDMB dataset. The reason why ICC has poor performance is that it only exploits temporal information that encodes temporal dependencies of mouse social behaviours to explore the variations of related frames. However, this approach may lead to suboptimal representation learning performance due to complex and diverse behavioural patterns in long videos. In contrast, our proposed SMC integrates both temporal and semantic information, which encodes the behavioural correlations and behaviour-specific characteristics, respectively. By exploiting the intra- and inter-information variations, our framework is able to learn more discriminative frame-wise representations.

We have also compared our entire framework, i.e., SMC-NCA with ICC using different settings of labelled data on our PDMB dataset. As presented in Tab. \ref{tab:comparison_sota_pdmb}, our approach achieves superior performance over ICC using only 10$\%$, 50$\%$ as well as 100$\%$ labelled data, thus demonstrating the effectiveness of our semi-supervised learning framework in modelling behavioural correlations of mice. Notably, the accuracy of our method varies across different datasets, with lower accuracy observed on the mouse behaviour dataset compared to human action datasets. This discrepancy may arise from several factors. First, mouse behaviours are often more complex and variable \cite{jiang2021multi,zhou2022cross}, with shorter durations and more frequent transitions between behaviours, leading to intricate behavioural correlations that are challenging to capture accurately. Second, our proposed method uses a temporal feature extractor originally designed for human action segmentation \cite{singhania2023c2f}. While effective for human actions, it may not adequately capture the unique features of mouse behaviour due to fundamental differences in movement patterns and action sequences.

Finally, to demonstrate the applicability of the proposed framework to behaviour phenotyping of the mice with Parkinson’s disease, we investigate the behavioural correlations of both MPTP-treated mice and their control strains, as shown in Fig. S2.  The findings reveal that MPTP-treated mice are more likely to perform 'approach' after 'circle', 'up' or 'walk$\_$away' compared to the control group ('other' is excluded). Besides, MPTP-treated mice tend to exhibit 'sniff' behaviour while the normal mice show a higher propensity towards 'walk$\_$away' after 'chase'.

\section{Conclusions and future work}
We have presented SMC-NCA, a novel semantic-guided multi-level contrast framework that aims to yield more discriminative frame-wise representations by fully exploiting complementary information from semantic and temporal entities. The proposed SMC aims to explore intra- and inter-information variations for unsupervised representation learning by integrating both semantic and temporal information. To achieve this goal, three types of dense negative pairs are explicitly constructed to facilitate contrastive learning. Our SMC provides a comprehensive solution that leverages both semantic and temporal features to enable effective unsupervised representation learning for action segmentation. To alleviate the over-segmentation problems in the semi-supervised setting with only a small amount of labelled data, the NCA module fully utilises spatial consistency between the neighbourhoods centered at different frames. Our framework outperforms the other state-of-the-art approaches on the three public datasets and our PDMB dataset, demonstrating the generalisation and versatility of our proposed framework for datasets of different domains. 
However, for datasets such as Breakfast, which features video-level activity labels, our multi-level contrast framework needs to be further developed to handle the similarity of activities. We will leave this exploration as future work.  In addition, our approach and insights may also apply to other video tasks, particularly those involving long untrimmed videos, such as temporal action detection. 

When the amount of labelled data is very limited, such as 5$\%$, the performance of the current contrast-classify framework in a semi-supervised setting can be potentially impacted by the quality of the pseudo labels. To address this, our future work will focus on enhancing the quality of pseudo labels to further improve representation learning. Additionally, we plan to develop more robust temporal and semantic feature extractors and devise methods to more effectively leverage the complementary information between them. By improving both the pseudo label generation process and the feature extraction techniques, we aim to achieve better performance even with minimal labelled data.

\ifCLASSOPTIONcaptionsoff
  \newpage
\fi

\bibliographystyle{IEEEtran}
\bibliography{egbib}

\clearpage
\onecolumn

\setcounter{table}{0}
\setcounter{algorithm}{0}
\setcounter{figure}{0}
\setcounter{equation}{0}
\setcounter{page}{1}
\renewcommand\thefigure{S\arabic{figure}}
\renewcommand\thetable{S\arabic{table}}
\renewcommand\thealgorithm{S\arabic{algorithm}}

\section*{Supplementary A}

\begin{figure*}[h]
\begin{center}
\includegraphics[width=16cm]{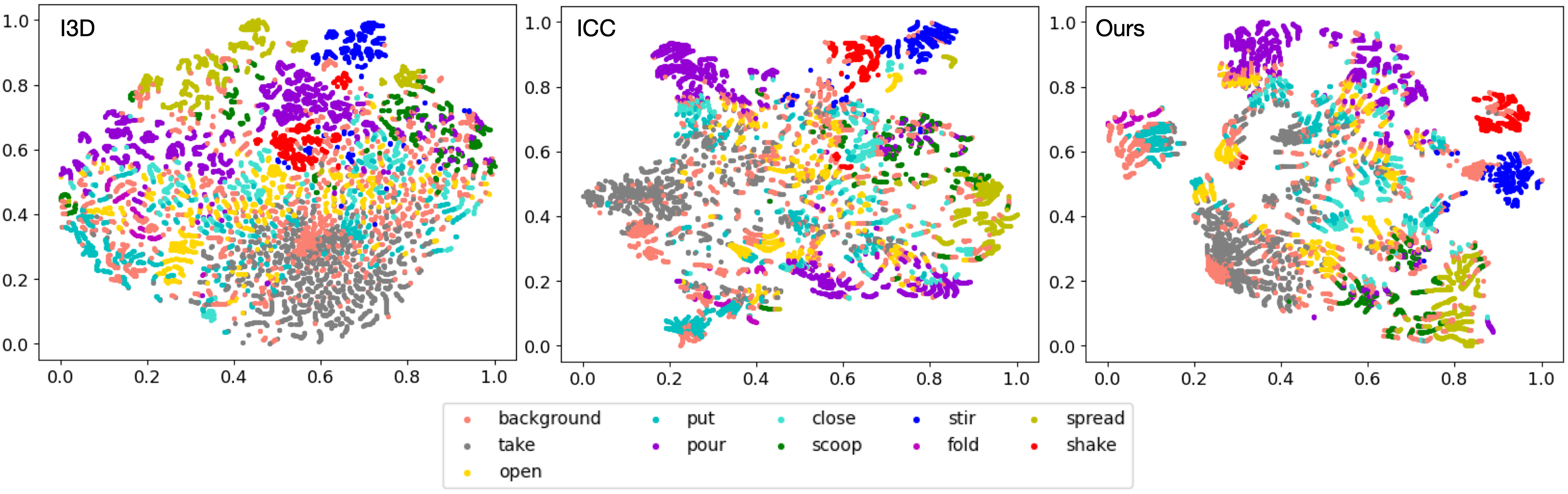}
\end{center}
\caption {t-SNE visualisation of the I3D feature and other features learned by ICC and our method. Each point represents an image frame. We show all behaviour classes (11) of the GTEA dataset in different colours.}
\label{fig:tsne_gtea}
\end{figure*}

\textbf{Additional qualitative results.} As shown in Fig. \ref{fig:tsne_gtea}, we visualise the I3D feature and other features learned by ICC and our method using t-Distributed Stochastic Neighbor Embedding (t-SNE) on GTEA datasets. Different colours represent different actions. Our approach leads to better separation of different classes, demonstrating the strong representation learning ability of our Semantic-guided Multi-level Contrast scheme.

\textbf{Compare SMC with other unsupervised representation learning methods}. In Tab. 3 (main paper) of the paper, we have compared our unsupervised method with that of ICC. Since ICC is the first work based on unsupervised representation learning for semi-supervised action segmentation, so far we could not find any other publications exploring unsupervised representation learning for this task. Thus, we attempt to compare our unsupervised method against a SOTA method \cite{dorkenwald2022scvrl} of unsupervised video representation learning, but it is not quite related to this task. We use the same positive and anchor representations as those of our work but the negative representation is obtained by temporal shuffling \cite{dorkenwald2022scvrl} of anchor representation. From Tab. \ref{tab:comparison_scvrl}, our method shows better performance.

\begin{table*}[h]
\centering
\caption{ Comparing Our SMC with SCVRL \cite{dorkenwald2022scvrl}. }
\label{tab:comparison_scvrl}
\begin{tabular}{m{1.2cm}<{\centering}|m{3.5cm}<{\centering}|m{2.8cm}<{\centering}m{0.6cm}m{0.6cm}}
\hline
Dataset & Method & F1@\{10, 25, 50\} \quad &Edit \quad &Acc\\
\hline
\hline
\multirow{2}{*}{50Salads} & SCVRL 
 \cite{dorkenwald2022scvrl} & 48.0\quad  43.5\quad  34.4\quad & 37.8 & 69.2\\
& SMC  & \textbf{58.1}\quad \textbf{54.0}\quad	\textbf{43.5}\qquad	&\textbf{46.3}\qquad	&\textbf{75.1}\\ 
\hline
\multirow{2}{*}{GTEA} & SCVRL \cite{dorkenwald2022scvrl} &72.5\quad 66.0\quad 48.6\quad &64.9\quad &71.2\\
& SMC  & \textbf{78.9}\quad	\textbf{74.3}\quad	\textbf{59.2}	&\textbf{73.0}	&\textbf{76.2} \\ 
\hline
\multirow{2}{*}{Breakfast} & SCVRL\cite{dorkenwald2022scvrl} &46.8\quad  41.9\quad  31.8\quad & 38.6\quad & 71.1\\
& SMC  &\textbf{59.7}\quad \textbf{55.4}\quad \textbf{42.8} & \textbf{52.7} & \textbf{72.1} \\ 
\hline
\end{tabular}
\end{table*}

\begin{table*}
\centering
\caption{Training hyperparameters learning rate (LR), weight-decay (WD), epochs (Eps.) and Batch Size (BS) over different datasets for unsupervised and semi-supervised learning.  }
\label{tab:hyperpara}
\begin{tabular}{l|l|l|l|l}
\hline
\multirow{2}{*}{Model}                                              & \qquad \quad Breakfast      &\qquad \quad 50Salads       &\qquad \quad  GTEA     &\qquad \quad  PDMB            \\\cline{2-5}
 & LR  \quad WD \quad Eps. \quad BS      & LR \quad WD \quad Eps.  \quad BS        & LR \quad WD  \quad Eps. \quad Bs   & LR \quad WD  \quad Eps. \quad Bs\\
 \hline
 \hline
 ($\textbf{T}$: $\textbf{S}$)  & le-3\quad 3e-3 \quad100 \quad50 & 1e-3 \quad1e-3 \quad100\quad 5 & 1e-3\quad 3e-4 \quad100 \quad4  & 1e-3\quad 3e-4 \quad100 \quad4 \\ 
\hline
 $\textbf{C}$  & le-2\quad 3e-3 \quad200 \quad50 & 1e-2 \quad1e-3 \quad400\quad 5 & 1e-2\quad 3e-4 \quad 400 \quad4   & 1e-2\quad 3e-4 \quad 400 \quad4\\ 
 ($\textbf{T}$:$\textbf{G}$:$\textbf{S}$)  & le-5\quad 3e-3 \quad 200 \quad 50  & 1e-5 \quad1e-3 \quad400\quad 5 & 1e-5\quad 3e-4 \quad 400 \quad 4  & 1e-5\quad 3e-4 \quad 400 \quad 4  \\ 
 
\hline
\end{tabular}
\end{table*}


\clearpage
\onecolumn

\section*{Supplementary B}
\begin{table*}[htbp]
\begin{center}
\caption{Ethogram of the observed behaviours \cite{jiang2021multi}. }
\label{tab:behaviour}
\begin{tabular}{p{2cm}|p{14cm}}
\toprule
Behaviour\quad & Description \\
\midrule
approach  & Moving toward another mouse in a straight line without obvious exploration.\\
chase & A following mouse attempts to maintain a close distance to another mouse while the latter is moving.\\
circle &Circling around own axis or chasing tail.\\
eat & Gnawing/eating food pellets held by the fore-paws.\\
clean &Washing the muzzle with fore-paws (including licking fore-paws) or grooming the fur or hind-paws by means of licking or chewing.\\
sniff & Sniff any body part of another mouse.\\
up & Exploring while standing in an upright posture.\\
walk away & Moving away from another mouse in a straight line without obvious exploration.\\
other & ehaviour other than defined in this ethogram, or when it is not visible what behaviour the mouse displays.\\
\bottomrule
\end{tabular}
\end{center}
\end{table*}

\textbf{Additional related works about temporal modelling of mouse behaviour.} In recent years, temporal dependencies among actions have also been investigated to facilitate mouse behaviour modelling. Jiang et al. \cite{jiang2018context} employed a Hidden Markov Model (HMM) to model the contextual relationship among adjacent mouse behaviours over time. Specifically, they represented each action clip as a set of feature vectors using spatial-temporal Segment Fisher Vectors (SFV), which were then treated as observed variables in the HMM. In addition, Jiang et al. \cite{jiang2021multi} proposed a deep graphic model to explore the temporal correlations of mouse social behaviours, which demonstrated the advantage of modelling behavioural correlations. However, these methods mainly focus on the correlations between the neighbouring behaviours, which is difficult to capture multi-scale temporal dependencies of mouse behaviours in long videos. Also, these methods usually require fully supervised data, which is obtained by manually annotating the exact temporal location of each behaviour occurring in all training videos. Such data collection is expensive, particularly in behavioural neuroscience \cite{pereira2020quantifying}, where datasets are usually complex and lab-specific. 

\begin{table*}[htb]
\centering
\caption{Component-wise analysis of the unsupervised representation learning framework with a linear classifier on the PDMB dataset.  }
\label{tab:ablation_unsuper_pdmb}
\begin{tabular}{m{5cm}<{\centering}|m{9.5cm}<{\centering}}
\hline
\multirow{2}{*}{Method}                                                    &PDMB        \\\cline{2-2}
  & F1@\{10, 25, 50\} \qquad Edit \qquad Acc   \\
 \hline
 \hline
$ \mathcal{L}_{ap}^{P}$($\textbf{M}_{in}$)+$ \mathcal{L}_{aa}^{N}$ & 39.0 \quad	33.9\quad	21.6\qquad	36.5\qquad37.7 \\

$ \mathcal{L}_{ap}^{P}$($\textbf{I}$)+$ \mathcal{L}_{aa}^{N}$ &\textbf{56.3} \quad	\textbf{53.6}\quad	\textbf{40.8}\qquad	\textbf{51.8}\qquad	\textbf{53.4} \\
\cline{2-2}
$ \mathcal{L}_{ap}^{P}$($\textbf{M}_{in}$)+$ \mathcal{L}_{ap}^{N}$ &38.0 \quad	34.2\quad	24.3\qquad	35.3\qquad	42.2\\

$ \mathcal{L}_{ap}^{P}$($\textbf{I}$)+$ \mathcal{L}_{ap}^{N}$ &\textbf{55.9} \quad	\textbf{53.5}\quad	\textbf{41.0}\qquad	\textbf{40.1}\qquad	\textbf{54.3} \quad \\
\hline
\multicolumn{2}{c}{\textit{ Constructing positive pairs by $\textbf{M}_{in}$ 
or $I$}}\\
\hline
$\mathcal{L}_{ap}^{P}$ + $ \mathcal{L}_{aa}^{N}$  &56.3 \quad	53.6\quad	40.8\qquad	51.8\qquad	53.4 \\

$\mathcal{L}_{ap}^{P}$ + $ \mathcal{L}_{ap}^{N}$  &55.9 \quad	53.5\quad	41.0\qquad	40.1\qquad	54.3 \quad \\

$\mathcal{L}_{ap}^{P}$ +  $ \mathcal{L}_{aa}^{N}$ + $ \mathcal{L}_{ap}^{N}$   & 58.5 \quad 56.6\quad	43.6\qquad	53.4\qquad	58.5 \\

$\mathcal{L}_{ap}^{P}$ +  $ \mathcal{L}_{aa}^{N}$ + $ \mathcal{L}_{ap}^{N}$ + $ \mathcal{L}_{pp}^{N}$  & \textbf{59.6} \quad	\textbf{58.0}\quad	\textbf{45.9}\qquad	\textbf{53.8}\qquad	\textbf{61.3} \\
\hline
\multicolumn{2}{c}{\textit{Comparing different negative pairs }}\\
\hline
w/o dynamic clustering & 59.6 \quad	58.0\quad	45.9\qquad	53.8\qquad	61.3  \\
w/ dynamic clustering & \textbf{61.1} \quad	\textbf{59.6}\quad	\textbf{47.2}\qquad	\textbf{55.5}\qquad	\textbf{62.5} \\
\hline
\multicolumn{2}{c}{\textit{Dynamic clustering facilitates contrastive learning  }}\\
\hline

\end{tabular}
\end{table*}

\begin{table*}
\centering
\caption{Performance of the NCA module on our PDMB dataset (10$\%$).}
\label{tab:ablatio_NCA_mouse}
\begin{tabular}{m{1.2cm}<{\centering}|m{2.5cm}<{\centering}|m{4.4cm}<{\centering}m{0.6cm}m{0.6cm}}      \hline
Dataset & Method & F1@\{10, 25, 50\} \quad &Edit \quad &Acc\\
\hline
\hline
\multirow{4}{*}{PDMB} & w/o $\mathcal{L}_{nca}$ &54.7\quad	48.7\quad	31.9	&45.0	&52.7 \\
& w/ $\mathcal{L}_{nca}$ & 62.0\quad	57.1\quad40.7	&52.7	&55.1 \\
& Gain  &\textbf{ 7.3} \quad	\textbf{ 8.4} \quad \textbf{8.8}	\qquad &\textbf{ 7.7}	&\textbf{ 2.4}\\ 
\hline
\end{tabular}
\end{table*}
\textbf{Mouse Social Behaviour Dataset.} Our Parkinson’s Disease Mouse Behaviour (PDMB) dataset was collected in collaboration with the biologists of Queen’s University Belfast of United Kingdom, for a study on motion recordings of mice with Parkinson’s disease (PD)  \cite{jiang2021multi}. The neurotoxin 1-methyl-4-phenyl-1,2,3,6-tetrahydropyridine (MPTP) is used as a model of PD, which has become an invaluable aid to produce experimental parkinsonism since its discovery in 1983  \cite{jackson2007protocol}. All experimental procedures were performed in accordance with the Guidance on the Operation of the Animals (Scientific Procedures) Act, 1986 (UK) and approved by the Queen’s University Belfast Animal Welfare and Ethical Review Body. We recorded videos for 3 groups of MPTP treated mice and 3 groups of control mice by using three synchronised Sony Action cameras (HDR-AS15) (one top-view and two side-view) with frame rate of 30 fps and 640*480 resolution. Each group consists of 6 annotated videos and all videos contain 9 behaviours (defined in Tab. \ref{tab:behaviour}) of two freely behaving mice. Different from the experiments of human action segmentation, the input features we use in experiments on this dataset are extracted from the pre-trained model from  \cite{zhou2022cross}, which encodes the social interactions of mice based on the pose information  \cite{zhou2021structured}. The whole dataset is evenly divided into training and testing datasets, and we select 10$\%$ or 50$\%$ of the videos from the training split for the labelled dataset $\mathcal{D}_{L}$.

\textbf{Evaluation of Representation Learning on the PDMB Dataset.} As shown in Tab. \ref{tab:ablation_unsuper_pdmb},  on the PDMB dataset, utilising $\textbf{M}_{in}$ would also lead to the generation of pseudo positive pairs, which would impede the efficacy of contrastive learning and consequently result in a significant performance drop. Besides, we achieve the best performance for all metrics when combining three types of negative pairs at the same time, where such combination brings gains of 7.9$\%$ and 7$\%$ in accuracy for the settings with only $ \mathcal{L}_{aa}^{N}$ and $ \mathcal{L}_{ap}^{N}$, respectively.

\textbf{Effect of the NCA Unit on the PDMB Dataset.} As shown in Tab. \ref{tab:ablatio_NCA_mouse}, with respect to behavioural correlation modelling of mice, we achieve a significant improvement of more than 7$\%$ in F1 and Edit scores on the PDMB dataset.

\begin{figure*}[htbp]
\begin{center}
\begin{tabular}{ccc}
\includegraphics[width=7cm]{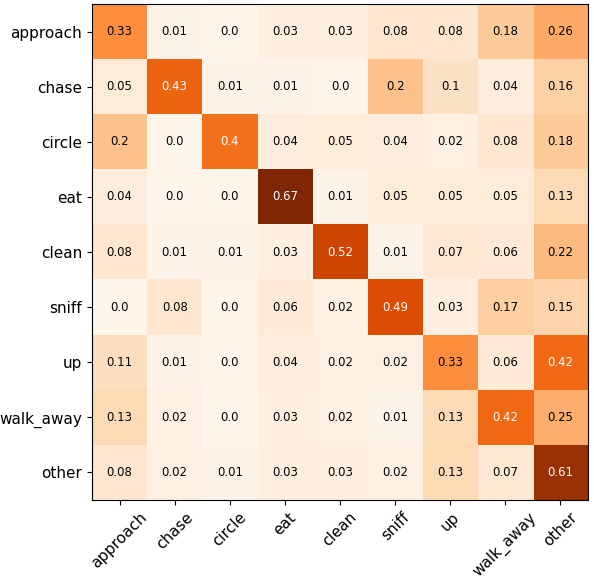}&
\hspace{-5mm}
\includegraphics[width=8.3cm]{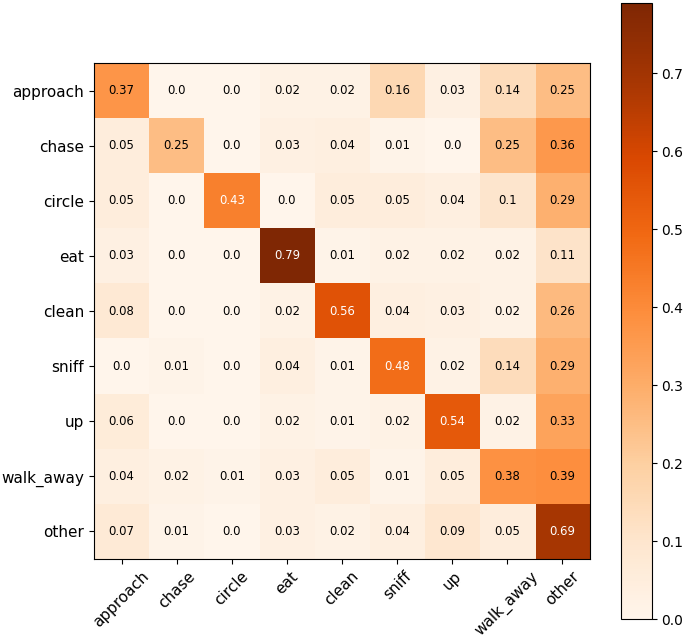}\\
(a) Parkinson’s Disease mice &(b) Normal mice
\end{tabular}
\end{center}
\caption[Occurrence frequency of neighbouring behaviours (20-frame interval) for Parkinson’s Disease and normal mice] {Occurrence frequency of neighbouring behaviours (20-frame interval) for Parkinson’s Disease and normal mice. Each cell contains the percentage of the occurrence of behaviour x (along rows) after behaviour y (along column) appears.}
\label{fig:occurrence_frequency_20frame}
\end{figure*}

Finally, to demonstrate the applicability of the proposed framework to behaviour phenotyping of the mice with Parkinson’s disease, we investigate the behavioural correlations of both MPTP treated mice and their control strains, as shown in Fig.  \ref{fig:occurrence_frequency_20frame}.  The findings reveal that MPTP treated mice are more likely to perform 'approach' after 'circle', 'up' or 'walk$\_$away' compared to the control group ('other' is excluded). Besides, MPTP treated mice tend to exhibit 'sniff' behaviour while the normal mice show a higher propensity towards 'walk$\_$away' after 'chase'.

\end{document}